%% file: acl2019.tex
\documentclass[11pt,a4paper]{article}
\usepackage[hyperref]{acl2019}
\usepackage{times}
\usepackage{latexsym}
\pdfoutput=1

\usepackage{amsmath}
\usepackage{amssymb}
\usepackage{ctable}
\usepackage{enumitem}
\usepackage{pifont}
\usepackage{makecell}
\usepackage{url}
\usepackage{graphicx}
\usepackage{footnote}
\usepackage{tablefootnote}
\DeclareMathOperator{\softmax}{\textbf{softmax}}

\makesavenoteenv{tabular}
\makesavenoteenv{table}
\graphicspath{{images/}}

\aclfinalcopy % Uncomment this line for the final submission
\def\aclpaperid{2009} %  Enter the acl Paper ID here

\newcommand{\xmark}{\ding{55}}%
\newcommand{\Ours}{SParC}
\newcommand{\syncon}{SyntaxSQL-con}
\newcommand{\syninp}{SyntaxSQL-sta}
\newcommand{\seqcon}{CD-Seq2Seq}

\newcommand*{\affaddr}[1]{#1} % No op here. Customize it for different styles.
\newcommand*{\affmark}[1][*]{\textsuperscript{#1}}
\newcommand*{\email}[1]{\texttt{#1}}
\newcommand{\newvec}[1]{\mathbf{#1}}

\newcommand{\vic}[1]{\textbf{\textcolor{magenta}{VL: #1}}}
\newcommand{\tz}[1]{\textbf{\textcolor{blue}{TS: #1}}}

\newcommand{\hide}[1]{}

\title{\Ours{}: Cross-Domain Semantic Parsing in Context}
\author{
Tao Yu\affmark[$\dagger$] 
\quad Rui Zhang\affmark[$\dagger$] 
\quad Michihiro Yasunaga\affmark[$\dagger$] 
\quad Yi Chern Tan\affmark[$\dagger$]
\quad Xi Victoria Lin\affmark[$\P$] 
\\{\bf Suyi Li\affmark[$\dagger$]
\quad Heyang Er\affmark[$\dagger$] 
\quad Irene Li\affmark[$\dagger$] 
\quad Bo Pang\affmark[$\dagger$]
\quad Tao Chen\affmark[$\dagger$] 
\quad Emily Ji\affmark[$\dagger$]} 
\\{\bf Shreya Dixit\affmark[$\dagger$]
\quad David Proctor\affmark[$\dagger$]
\quad Sungrok Shim\affmark[$\dagger$] 
\quad Jonathan Kraft\affmark[$\dagger$]} 
\\{\bf Vincent Zhang\affmark[$\dagger$]
\quad Caiming Xiong\affmark[$\P$]
\quad Richard Socher\affmark[$\P$]
\quad Dragomir Radev\affmark[$\dagger$]}\\
\affaddr{\affmark[$\dagger$]Department of Computer Science, Yale University}\\
\affaddr{\affmark[$\P$]Salesforce Research}\\
\email{\{tao.yu,\,r.zhang,\,michihiro.yasunaga,\,dragomir.radev\}@yale.edu}\\
\email{\{xilin,\,cxiong,\,rsocher\}@salesforce.com}
}

\date{}

\begin{document}
\maketitle
\begin{abstract}
  
We present \Ours{}, a dataset for cross-domain \textbf{S}emantic \textbf{Par}sing in \textbf{C}ontext.
% We present \Ours{}, a context-dependent semantic parsing challenge dataset for building cross-domain interactive semantic parsing systems.
% It consists of 12k+ questions annotated with SQL queries, obtained from 4,298 unique interactions inquiring 200 complex databases from 138 different domains.
It consists of 4,298 coherent question sequences (12k+ individual questions annotated with SQL queries), obtained from controlled user interactions with 200 complex databases over 138 domains.
% We analyze the dataset in depth and show that \Ours{} introduces new challenges not found in prior datasets: 
% (1) questions in \Ours{} cover wider semantic meaning and more diverse semantic changes in context;
% (2) the database split in \Ours{} requires systems to generalize to unseen domains.
We provide an in-depth analysis of \Ours{} and show that it introduces new challenges compared to existing datasets. 
\Ours{}
(1) demonstrates complex contextual dependencies,
% (2) have greater diversity in semantic coverage, and
(2) has greater semantic diversity, and
(3) requires generalization to new domains due to its cross-domain nature and the unseen databases at test time.
% (2) the large number of domains and cross-domain evaluation require generalization to unseen domains.
We experiment with two state-of-the-art text-to-SQL models adapted to the context-dependent, cross-domain setup.
The best model obtains an exact set match accuracy of 20.2\% over all questions and less than 10\% over all interaction sequences, indicating that the cross-domain setting and the contextual phenomena of the dataset present significant challenges for future research.
The dataset, baselines, and leaderboard are released at \url{https://yale-lily.github.io/sparc}.

\end{abstract}

\input{introduction.tex}
\input{related_work.tex}
\input{data_collection.tex}
\input{data_statistics.tex}
\input{methods.tex}
\input{results.tex}
\input{conclusion.tex}
\input{acknowledgements.tex}

\bibliography{acl2019}
\bibliographystyle{acl_natbib}

\appendix
\input{appendix.tex}

\end{document}

%% file: introduction.tex
\section{Introduction}
% \vic{Tentatively changing the title to help flesh out the story.}
% \vic{High-level: critics of the ATIS dataset seems too harsh at several places throughout the paper. Given that most people in the semantic parsing domain had worked on this dataset and it inspired a lot of models that are still being used today, some of the critics could sound controversial to people (criticizing the dataset also criticizes those modeling work). In table~\ref{tb:data_stats}, \Ours{} is not orders of magnitude larger than ATIS, which will also weaken some of these arguments. Rewriting suggestion: focus on what the new dataset \Ours{} adds to the field and describe the shortcomings of ATIS succinctly. (I can help w/ the rewriting. No need to worry if there are other tasks of higher priority.)}

% \vic{Please fill in the author list and switch to the camera-ready template. We have a long author list and this enables us to accurately estimate the space left in the paper.}
% \tz{One sentence explaining ``why database?''}
Querying a relational database is often challenging and a natural language interface has long been regarded by many as the most powerful database interface~\cite{popescu2003towards,Bertomeu06,li2014constructing}. 
The problem of mapping a natural language utterance into executable SQL queries (text-to-SQL) has attracted increasing attention from the semantic parsing community by virtue of a continuous effort of dataset creation~\cite{zelle96,Iyyer17,Zhong2017,cathy18,Yu18} and the modeling innovation that follows it~\cite{Xu2017,2018executionguided,Yu&al.18.emnlp.syntax,shi2018incsql}. 
% \tz{Suggest rewrite (more positive tone in describing the prior papers): talk about the different goals in dataset creation, some mainly support question answering, and some support interactive data exploration.}
\begin{figure}[t!]
    % \vspace{-1.5mm}\hspace{-1mm}
    \centering
    \includegraphics[width=0.48\textwidth]{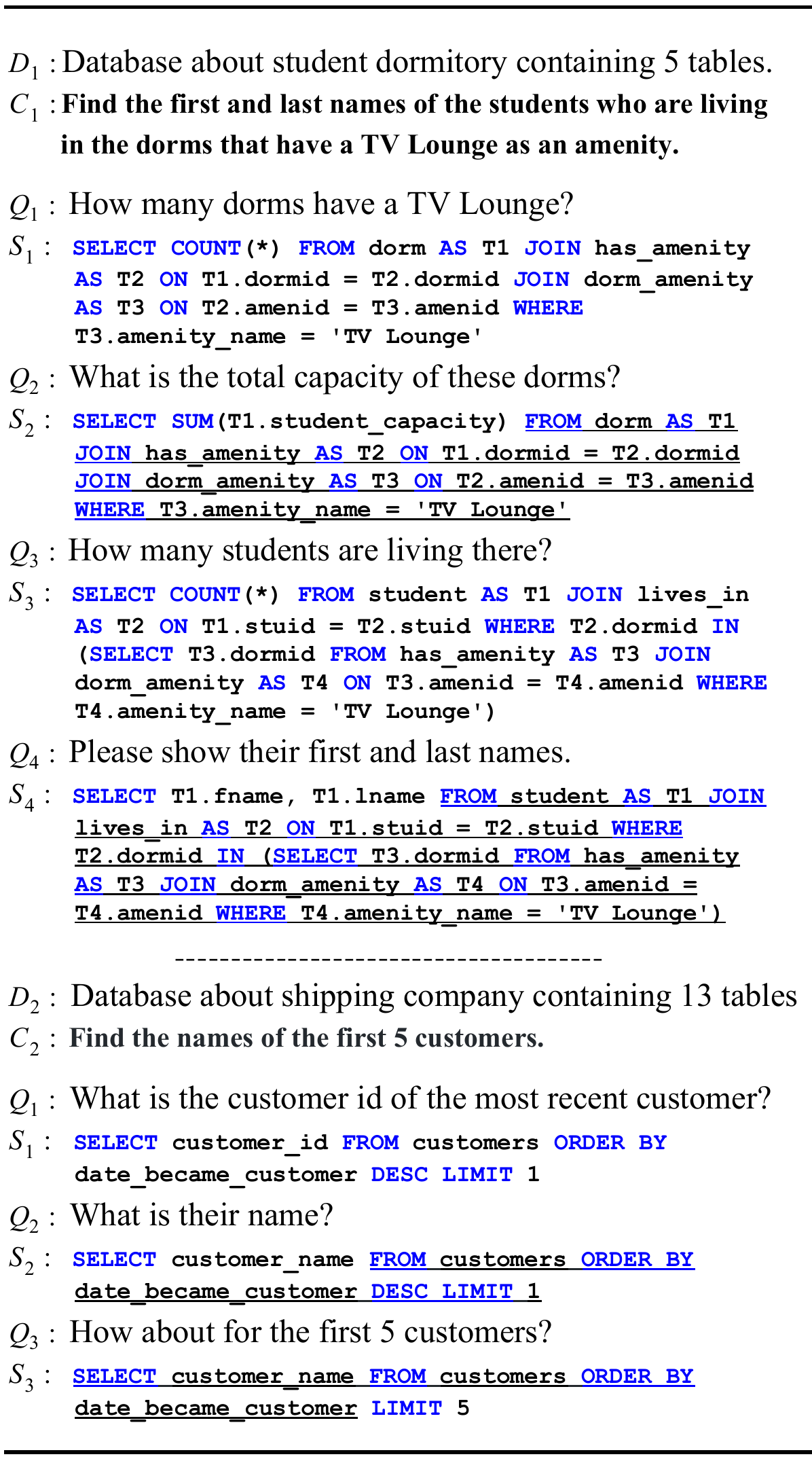}
    \vspace{-2mm}
    \caption{Two question sequences from the \Ours{} dataset. Questions ($Q_i$) in each sequence query a database ($D_m$), obtaining information sufficient to complete the interaction goal ($C_m$). Each question is annotated with a corresponding SQL query ($S_i$). 
    SQL segments from the interaction context are underlined.
    % SQL token sequences inherited from previous turns are underlined.
    }
\label{fig:task}
\vspace{-6.54mm}
\end{figure}

While most of these work focus on precisely mapping stand-alone utterances to SQL queries, generating SQL queries in a context-dependent scenario~\cite{Miller96,Zettlemoyer09,Suhr:18context} has been studied less often. The most prominent context-dependent text-to-SQL benchmark is ATIS\footnote{A subset of ATIS is also frequently used in context-independent semantic parsing research~\cite{D07-1071,dong16}.}, which is set in the flight-booking domain and contains only one database~\cite{Hemphill90,Dahl94}.

% Previous studies show that in practice, the information needed to correctly answer a user query often goes beyond a single stand-alone utterance~\cite{Kato2004HandlingIA,Chai04,Bertomeu06}. 

In a real-world setting, users tend to ask a sequence of thematically related questions to learn about a particular topic or to achieve a complex goal. Previous studies have shown that by allowing questions to be constructed sequentially, users can explore the data in a more flexible manner, which reduces their cognitive burden~\cite{Hale06,Levy08,Frank13,Iyyer17} and increases their involvement when interacting with the system.
The phrasing of such questions depends heavily on the interaction history~\cite{Kato2004HandlingIA,Chai04,Bertomeu06}.
The users may explicitly refer to or omit previously mentioned entities and constraints, and may introduce refinements, additions or substitutions to what has already been said (Figure~\ref{fig:task}).
% As shown in Figure~\ref{fig:task}, the user may either explicitly refer to or omit previously mentioned constraints and entities (e.g. $\mathcal{Q}_{11}\rightarrow\mathcal{Q}_{12}$), $\mathcal{Q}_{21}\rightarrow\mathcal{Q}_{22}$). 
This requires a practical text-to-SQL system to effectively process context information to synthesize the correct SQL logic.
% On the other hand, the SQL queries corresponding to such thematically related questions often share common segments. 
% The SQL queries corresponding to such thematically related questions often have significant overlap. 
% Consider the two interaction sequences in Figure~\ref{fig:task}. In the first one the user is researching on groups of customers of the shipping company registered at different time, and in the second one the user is querying 

% To address the need for high-quality context-dependent text-to-SQL data, 
To enable modeling advances in context-dependent semantic parsing, we introduce \Ours{} (cross-domain \textbf{S}emantic \textbf{Par}sing in \textbf{C}ontext), an expert-labeled dataset which contains 4,298 coherent question sequences (12k+ questions paired with SQL queries) querying 200 complex databases in 138 different domains. 
% we introduce the Context-Dependent Sequential Question to SQL (\Ours{}) corpus and task.
The dataset is built on top of Spider\footnote{The data is available at \url{https://yale-lily.github.io/spider}.}, the largest cross-domain context-independent text-to-SQL dataset available in the field~\cite{Yu18emnlp}. The large number of domains provide rich contextual phenomena and thematic relations between the questions, which general-purpose natural language interfaces to databases have to address. In addition, it enables us to test the generalization of the trained systems to unseen databases and domains.

% To obtain question sequences over the Spider databases, we asked 15 college students with SQL background to explore the database content and produce sequences of simple but related questions during their exploration. 
We asked 15 college students with SQL experience to come up with question sequences over the Spider databases (\S~\ref{sec:data_collection}). 
% The dataset was labeled by 15 college students with SQL background.
Questions in the original Spider dataset were used as guidance to the students for constructing meaningful interactions: each sequence is based on a question in Spider and the student has to ask inter-related questions to obtain information that answers the Spider question. At the same time, the students are encouraged to come up with related questions which do not directly contribute to the Spider question so as to increase data diversity. The questions were subsequently translated to complex SQL queries by the same student. Similar to Spider, the SQL Queries in \Ours{} cover complex syntactic structures and most common SQL keywords.
% that have final semantics equal to the complicated questions in Spider. \tz{this is repeating last sentence in the last paragraph.}
% , which shows its wide diversity in semantic coverage. \tz{SQL structures = semantic diversity?}
% Also, questions in Spider and \Ours{} both query 200 complex databases with multiple tables in 138 different domains.
% All questions and SQL queries are annotated by 15 college students with a background in databases.

We split the dataset such that a database appears in only one of the train, development and test sets.
% which requires systems to generalize to not only new SQL queries\vic{new SQL queries or new questions?} but also new database schemas.
% In this way, it is possible to develop domain-free semantic parsers without any new data collection and model retraining steps.
% Therefore, comparing with ATIS \tz{a little abrupt to mention ATIS, needs transition}, 
% As a result, to do well on \Ours{} task, a model needs to
% \begin{enumerate}[topsep=6pt,itemsep=-1ex,label=(\alph*)]
%     \item handle complex SQL representations; 
%     \item handle a diverse set of contextual phenomena between questions;
%     \item be able to generalize across different databases.
% \end{enumerate}
We provide detailed data analysis to show the richness of \Ours{} in terms of semantics, contextual phenomena and thematic relations (\S~\ref{sec:data_stats}).
We also experiment with two competitive baseline models to assess the difficulty of \Ours{} (\S~\ref{sec:baselines}).
The best model achieves only 20.2\% exact set matching accuracy\footnote{Exact string match ignores ordering discrepancies of SQL components whose order does not matter. Exact set matching is able to consider ordering issues in SQL evaluation. See more evaluation details in section \ref{sec:eval}.} on all questions, and demonstrates a decrease in exact set matching accuracy from 38.6\% for questions in turn 1 to 1.1\% for questions in turns 4 and higher (\S~\ref{sec:results}).
This suggests that there is plenty of room for advancement in modeling and learning on the \Ours{} dataset.

%% file: related_work.tex
\section{Related Work}

\input{tables/data_compare.tex}

% \tz{This paragraph reads redundant} Some related work and datasets include contextual-independent \cite{Zettlemoyer05} and contextual-dependent \cite{Suhr:18context} semantic parsing, and sequential question representation learning \cite{long16,Iyyer17}.

\paragraph{Context-independent semantic parsing} Early studies in semantic parsing \cite{Zettlemoyer05,artzi13,Berant14,li2014constructing,pasupat2015compositional,dong16,iyer17} were based on small and single-domain datasets such as ATIS \cite{Hemphill90,Dahl94} and GeoQuery \cite{zelle96}.
%These datasets have a limited number of labeled logic forms or SQL queries. In order to expand the size of these datasets and apply neural network approaches, each logic form or SQL query has about 4-10 paraphrases for the natural language input.
Recently, an increasing number of neural approaches \cite{Zhong2017,Xu2017,Yu18,dong18,Yu&al.18.emnlp.syntax} have started to use large and cross-domain text-to-SQL datasets such as WikiSQL \cite{Zhong2017} and Spider \cite{Yu18emnlp}.
Most of them focus on converting stand-alone natural language questions to executable queries. Table \ref{tb:data_compare} compares \Ours{} with other semantic parsing datasets.

\paragraph{Context-dependent semantic parsing with SQL labels} Only a few datasets have been constructed for the purpose of mapping context-dependent questions to structured queries.
\citet{Hemphill90,Dahl94} collected the contextualized version of ATIS that includes series of questions from users interacting with a flight database. 
%\citet{Suhr:18context} study with the contextual version of ATIS, and introduce a context-dependent model mapping questions in context to SQL queries.
Adopted by several works later on \cite{Miller96,Zettlemoyer09,Suhr:18context}, ATIS has only a single domain for flight planning which limits the possible SQL logic it contains.
In contrast to ATIS, \Ours{} consists of a large number of complex SQL queries (with most SQL syntax components) inquiring 200 databases in 138 different domains, which contributes to its % semantic and contextual diversity.
diversity in query semantics and contextual dependencies.
% Most importantly, 
Similar to Spider, % and \citet{cathy18},  
the databases in the train, development and test sets of \Ours{} do not overlap.
%Another similar work is CONCODE \cite{Iyer19}, which is for generating class member functions given English documentation and programmatic context.

\paragraph{Context-dependent semantic parsing with denotations} 
%\vic{SCONE is considered as a semantic parsing dataset by the field. It is a weakly supervised semantic parsing dataset where only the correct execution result is labeled.}
Some datasets used in recovering context-dependent meaning (including SCONE \cite{long16} and SequentialQA \cite{Iyyer17}) contain no logical form annotations but only denotation~\cite{Berant14} instead.
SCONE \cite{long16} contains some instructions in limited domains such as chemistry experiments.
The formal representations in the dataset are world states representing state changes after each instruction instead of programs or logical forms.
SequentialQA \cite{Iyyer17} was created by asking crowd workers to decompose some complicated questions in WikiTableQuestions \cite{pasupat2015compositional} into sequences of inner-related simple questions.
As shown in Table \ref{tb:data_compare}, neither of the two datasets were annotated with query labels.
Thus, to make the tasks feasible, SCONE \cite{long16} and SequentialQA \cite{Iyyer17} exclude many questions with rich semantic and contextual types.
For example, \cite{Iyyer17} requires that the answers to the questions in SequentialQA must appear in the table, and most of them can be solved by simple SQL queries with \texttt{SELECT} and \texttt{WHERE} clauses.
Such direct mapping without formal query labels becomes unfeasible for complex questions.
Furthermore, SequentialQA contains questions based only on a single Wikipedia tables at a time.
In contrast, \Ours{} contains 200 significantly larger databases, and complex query labels with all common SQL key components.
This requires a system developed for \Ours{} to handle % more complex 
information needed over larger databases in different domains.

\paragraph{Conversational QA and dialogue system} 
Language understanding in context is also studied for dialogue and question answering systems. % over text.
% Studies 
The development in dialogue \cite{Henderson2014,mrksic17,zhong2018global} uses predefined ontology and slot-value pairs with limited natural language meaning representation, whereas we focus on % unconstrained 
general SQL queries that enable more powerful semantic meaning representation.
% mention task-oriented dilogue datasets dstc 2
Recently, some % datasets in question answering dialogue for information-gathering 
conversational question answering datasets have been introduced, such as QuAC \cite{Eunsol18} and CoQA \cite{Reddy18}.
They differ from \Ours{} in that the answers are free-form text instead of SQL queries. % based on databases.
On the other hand, \citet{Kato2004HandlingIA,Chai04,Bertomeu06} conduct early studies % and findings 
of the contextual phenomena and thematic relations in database dialogue/QA systems, which we use as references when constructing \Ours{}.

%% file: tables/data_compare.tex
\iffalse %%%%%%%%%%%
\begin{table*}[ht!]
\centering
\scalebox{0.78}{
\begin{tabular}{ccccc}
\Xhline{2\arrayrulewidth}
Dataset & Context & Resource & Annotation & Cross-domain\\\hline
\textbf{\Ours{}} & \checkmark & database & SQL & \checkmark\\
ATIS \cite{Hemphill90,Dahl94} & \checkmark & database & SQL & \xmark\\\hline
Spider \cite{Yu&al.18.emnlp.corpus} & \xmark  & database & SQL & \checkmark\\
WikiSQL \cite{Zhong2017} & \xmark  & table & SQL & \checkmark\\
GeoQuery \cite{zelle96} & \xmark  & database & SQL & \xmark\\\hline
SequentialQA \cite{Iyyer17} & \checkmark & table & answer & \checkmark\\
SCONE \cite{long16} & \checkmark & environment & answer & \checkmark \\\hline
CoQA \cite{Reddy18} & \checkmark & article & answer & \checkmark \\
QuAC \cite{Eunsol18} & \checkmark & article & answer & \checkmark\\
\Xhline{2\arrayrulewidth}
\end{tabular}}
\caption{A survey of several context-dependent semantic parsing and question answering datasets. add some descriptions instead of just numbers. add table number per domain. provide dialogue length distribution. mention many correferences in SequentialQA not very diverse, analyze SequentialQA's thematic relations. ATIS contains many repeated joins and nested queries but without many other sql components.} 
\label{tb:data_compare}
\end{table*}
\fi %%%%%%%%%%%

\begin{table*}[ht!]
\centering
\scalebox{0.90}{
\begin{tabular}{ccccc}
\Xhline{2\arrayrulewidth}
Dataset & Context & Resource & Annotation & Cross-domain\\\hline
\textbf{\Ours{}} & \checkmark & database & SQL & \checkmark\\
ATIS \cite{Hemphill90,Dahl94} & \checkmark & database & SQL & \xmark\\\hline
Spider \cite{Yu18emnlp} & \xmark  & database & SQL & \checkmark\\
WikiSQL \cite{Zhong2017} & \xmark  & table & SQL & \checkmark\\
GeoQuery \cite{zelle96} & \xmark  & database & SQL & \xmark\\\hline
SequentialQA \cite{Iyyer17} & \checkmark & table & denotation & \checkmark\\
SCONE \cite{long16} & \checkmark & environment & denotation & \xmark \\\hline
%CoQA \cite{Reddy18} & \checkmark & article & free-text & \checkmark \\
%QuAC \cite{Eunsol18} & \checkmark & article & free-text & \checkmark\\\hline
%DSTC2 \cite{Henderson2014} & \checkmark & database & dialog action & \xmark\\
%MultiWOZ \cite{Budzianowski2018MultiWOZA} & \checkmark & database & dialog action & \checkmark \\
\Xhline{2\arrayrulewidth}
\end{tabular}}
\caption{Comparison of \Ours{} with existing semantic parsing datasets.} %\vic{Need to rename the Target column. Consider removing the last four rows of this table. In general, QA and dialogue datasets are not count as semantic parsing datasets.}
\label{tb:data_compare}
\end{table*}

%% file: data_collection.tex
\section{Data Collection}
\label{sec:data_collection}

We create the \Ours{} dataset in four stages: selecting interaction goals,
% \vic{``final query goal'' is a strange term, started talking about ``final query goal'' before defining ``final query''. Better term? ``interaction goal''?}
creating questions, annotating SQL representations, and reviewing. % the data.

\input{tables/examples.tex}

\paragraph{Interaction goal selection}
To ensure thematic relevance within each question sequence, we use questions in the original Spider dataset as the thematic guidance for constructing meaningful query interactions, i.e. the interaction goal. Each sequence is based on a question in Spider and the annotator has to ask inter-related questions to obtain the information 
% that completes 
demanded by the interaction goal (detailed in the next section). 
All questions in Spider were stand-alone questions written by 11 college students with SQL background after they had explored the database content, and
% . While some of the questions turned out to be very complicated since the students were forced to express their objectives in a single turn, 
the question intent conveyed is likely to naturally arise in real-life query scenarios.
%, which simulates real-life general-purpose query scenarios. 
We selected all Spider examples classified as medium, hard, and extra hard, as it is in general hard to establish context for easy questions. In order to study more diverse information needs, we also included some easy examples (end up with using 12.9\% of the easy examples in Spider).
% \footnote{Already defined in the Spider task.}.
% The SQL answers of these questions include multiple SQL components such as \texttt{WHERE}, \texttt{ORDER BY}, \texttt{GROUP BY}, \texttt{EXCEPT}, and nested queries.
As a result, 4,437 questions were selected as the interaction goals for 200 databases.

\hide{
We used questions in Spider % \cite{Yu18emnlp} 
as users' final complex information requests.
Spider is the largest complex and cross-domain semantic parsing and text-to-SQL dataset consisting of 10k+ complex questions about a number of databases.
All questions in Spider were written by 11 college students with SQL backgrounds after they explored the content of the database, which simulates real-life general-purpose query scenarios. 
Using the Spider complex questions as final interaction goals avoids the repeated work of finding new query goals.
Although each complex question may be an unnatural utterance, this is because the complex informational demands are expressed in a single question turn. 
Even so, the question intent conveyed is likely to naturally arise in queries of databases, especially in a dialogue-based sequential setting after exploration of the databases. (we asked the annotators of \Ours{} to similarly explore the databases and ask more new questions, more details in next paragraph).
We selected all examples that are classified as medium, hard, and extra hard (in total \vic{XX} examples).
% \footnote{Already defined in the Spider task.}.
The SQL answers of these questions include multiple SQL components such as \texttt{WHERE}, \texttt{ORDER BY}, \texttt{GROUP BY}, \texttt{EXCEPT}, and nested queries.
In order to study more diverse information needs, we also included some easy examples.
Finally, 4,437 questions were selected as the interaction goals for 200 databases.
}

\paragraph{Question creation}
15 college students with SQL experience were asked to come up with sequences of inter-related questions to obtain the information demanded by the interaction goals\footnote{The students were asked to spend time exploring the database using a database visualization tool powered by Sqlite Web \url{https://github.com/coleifer/sqlite-web} so as to create a diverse set of thematic relations between the questions.}.  Previous work~\cite{Bertomeu06} has characterized different thematic relations between the utterances in a database QA system: \textbf{\textit{refinement}}, \textbf{\textit{theme-entity}}, \textbf{\textit{theme-property}}, and \textbf{\textit{answer refinement/theme}}\footnote{We group \textit{answer refinement} and \textit{answer theme}, the two thematic relations holding between a question and a previous answer as defined in \citet{Bertomeu06}, into a single \textit{answer refinement/theme} type.}, as shown in Table~\ref{tb:examples}. 
% \vic{Is the last relation name a typo? It is very hard to understand what it means based on its surface form. Should it be ``theme-refinement'' or something?} 
We show these definitions to the students prior to question creation to help them come up with % ``inter-related'' questions.  
context-dependent questions.
% Notice that we 
We also encourage the formulation of questions that are thematically related to but do not directly contribute to answering the goal question (e.g. $Q_2$ in the first example and $Q_1$ in the second example in Figure~\ref{fig:task}. See more examples in Appendix as well).
The students do not simply decompose the complex query.
Instead, they often explore the data content first and even change their querying focuses. Therefore, all interactive query information in \Ours{} could not be acquired by a single complex SQL query.

We divide the goals evenly among the students and each interaction goal is annotated by one student\footnote{The most productive student annotated 13.1\% of the goals and the least productive student annotated close to 2\%.}.
We enforce each question sequence to contain at least two questions, and the interaction terminates when the student has obtained enough information to answer the goal question.

\paragraph{SQL annotation}
After creating the questions, each annotator was asked to translate their own questions to SQL queries.
%\vic{Do the question writers answer their own questions? Or a different group of students answer them?} They could refer to the SQL answers to the original complex questions in the Spider dataset.
All SQL queries were executed on Sqlite Web to ensure correctness.
To make our evaluation more robust, the same annotation protocol as Spider \cite{Yu18emnlp} was adopted such that all annotators chose the same SQL query pattern when multiple equivalent queries were possible.
%\vic{Need to ensure the paper is self-contained, hence need to reiterate the annotation protcol again. If the protocol is too long to be included, make a citation to the Spider paper here.} 

\paragraph{Data review and post-process}
We asked students who are native English speakers to review the annotated data. Each example was reviewed at least once.
The students corrected any grammar errors and rephrased the question in a more natural way if necessary.
They also checked if the questions in each sequence were related and the SQL answers matched the semantic meaning of the question.
After that, another group of students ran all annotated SQL queries to make sure they were executable.
Furthermore, they used the SQL parser\footnote{\url{https://github.com/taoyds/spider/blob/master/process_sql.py}} from Spider to parse all the SQL labels to make sure all queries follow the annotation protocol.
Finally, the most experienced annotator conducted a final review on all question-SQL pairs. 
139 question sequences were discarded in this final step due to poor question quality or wrong SQL annotations
%\vic{``poor data quality'' and ``wrong annotations'' are both vague expressions and seem to mean the same thing. State the exact criteria used for discarding the data in the final review here}.
%\vic{How many rounds of review were conducted? On average how many students reviewed the same sequence?}

%% file: tables/examples.tex
\begin{table*}[ht!]
\centering
\scalebox{0.78}{
\begin{tabular}{p{28mm}p{60mm}p{75mm}cc}
\Xhline{2\arrayrulewidth}
Thematic relation & Description & Example & Percentage  \\ \hline
%the subset, superset or disjoint set of
Refinement (constraint refinement) & The current question asks for the same type of entity as a previous question with a different constraint. & Prev\_Q: Which \textbf{major} has \underline{the fewest students}? \newline Cur\_Q: What is \underline{the most popular one}? & 33.8\%\\ \hline
Theme-entity (topic exploration) & The current question asks for other properties about the same entity as a previous question. & Prev\_Q: What is \textit{the capacity} of \textbf{Anonymous Donor Hall}? \newline Cur\_Q: List \textit{all of the amenities} which \textbf{it} has. & 48.4\% \\ \hline
Theme-property (participant shift) & The current question asks for the same property about another entity. & Prev\_Q: Tell me the \textit{rating} of \textbf{the episode named ``Double Down"}. \newline Cur\_Q: How about for \textbf{``Keepers"}? & 9.7\% \\ 
\specialrule{.1em}{.05em}{.05em} 
% \hline
Answer refinement/theme (answer exploration) & The current question asks for a subset of the entities given in a previous answer or asks about a specific entity introduced in a previous answer. % the entities in the answer set of previous question.  
& Prev\_Q: Please list all the different \textbf{department} \textit{names}. \newline Cur\_Q: What is the \textit{average salary} of all instructors in the \textbf{Statistics department}? & 8.1\% \\
% \Xhline{2\arrayrulewidth}
\bottomrule
\end{tabular}}
\caption{Thematic relations between questions in a database QA system defined by \citet{Bertomeu06}. The first three relations hold between a question and a previous question and the last relation holds between a question and a previous answer. We manually classified 102 examples in \Ours{} into one or more of them and show the distribution. The entities (\textbf{bold}), properties (\textit{italics}) and constraints (\underline{underlined}) are highlighted in each question.} 
\label{tb:examples}
\end{table*}

%% file: data_statistics.tex
\section{Data Statistics and Analysis}
\label{sec:data_stats}

%\vic{Revision needed (high priority) -- be very focused on describing the characteristics and contributions of \Ours{}. Most readers probably haven't seen the multi-turn ATIS dataset but should be able to immediately absorb the differences of these two, especially after reading Table~\ref{tb:data_stats}. Hence giving extensive description of ATIS is secondary. The only point regarding ATIS which should be carefully addressed is its higher \# of turns.}
\input{tables/data_stats.tex}

We compute the statistics of \Ours{} and conduct a through data analysis focusing on its contextual dependencies, semantic coverage and cross-domain property. Throughout this section, we compare \Ours{} to ATIS~\cite{Hemphill90,Dahl94}, the most widely used context-dependent text-to-SQL dataset in the field. 
% In comparison, \Ours{} is significantly different as it (1) contains much more complex context dependencies and changes between questions, (2) has greater semantic coverage and more diverse thematic relations, and (3) has a cross-domain task setting. 
In comparison, \Ours{} is significantly different as it (1) contains % much more complex context dependencies and more diverse thematic relations, 
mode complex contextual dependencies, (2) has greater semantic coverage, and (3) adopts a cross-domain task setting,
% This makes it the only large complex cross-domain and context-dependent text-to-SQL dataset.
which make it a new and challenging cross-domain context-dependent text-to-SQL dataset.
% We analyze the semantic complexity and overlap of questions in interactions, report the distribution of SQL components and of different types of thematic relations, and describe the database split statistics.

\paragraph{Data statistics}

Table \ref{tb:data_stats} summarizes the statistics of \Ours{} and ATIS.
\Ours{} contains 4,298 unique question sequences, 200 complex databases in 138 different domains, with 12k+ questions annotated with SQL queries. The number of sequences in ATIS is significantly smaller, but it contains a comparable number of individual questions since it has a higher number of turns per sequence\footnote{The ATIS dataset is collected under the Wizard-of-Oz setting~\cite{Bertomeu06} (like a task-oriented sequential question answering task). Each user interaction is guided by an abstract, high-level goal such as ``plan a trip from city A to city B, stop in another city on the way''. %and the user needs to conduct a multi-turn conversation to complete the goal. 
The domain by its nature requires the user to express multiple constraints in separate utterances and the user is intrinsically motivated to interact with the system until the booking is successful. In contrast, the interaction goals formed by Spider questions are for open-domain and general-purpose database querying, which tend to be more specific and can often be stated in a smaller number of turns. We believe these differences contribute to the shorter average question sequence length of \Ours{} compared to that of ATIS.}.
On the other hand, \Ours{} has overcome the domain limitation of ATIS by covering 200 different databases and has a significantly larger natural language vocabulary. 
% \vic{Evaluated by SQL coverage, ...}

\begin{figure}[!t]
    \centering
    \includegraphics[width=0.4\textwidth]{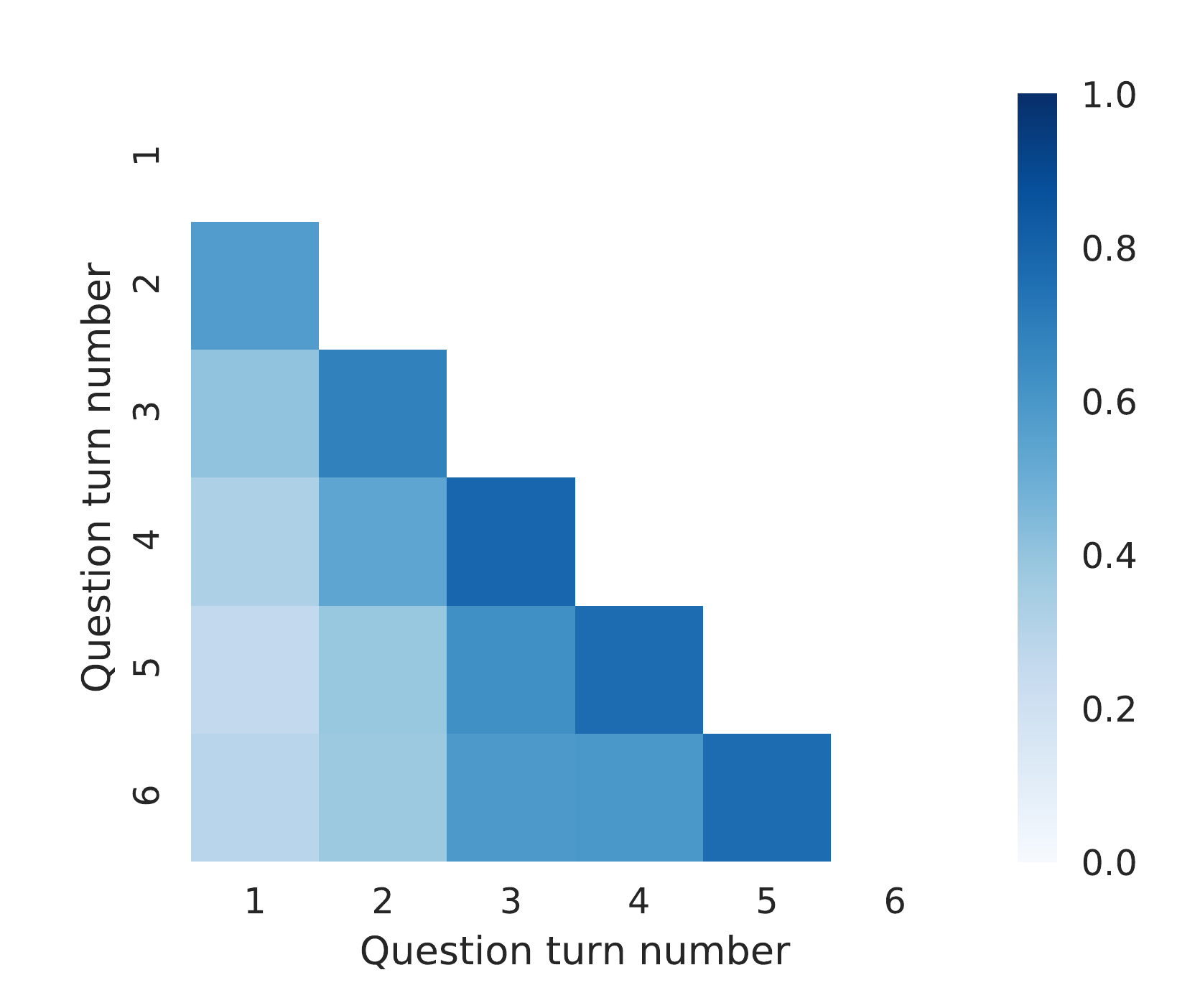}
    \vspace{-2mm}
    \caption{The heatmap shows the percentage of SQL token overlap between questions in different turns. Token overlap is greater between questions that are closer to each other and the degree of overlap increases as interaction proceeds. % which demonstrates the prevalence of contextual dependencies in \Ours{}. 
    Most questions have dependencies that span 3 or fewer turns.}
\label{fig:sql_reuse}
\vspace{-2mm}
\end{figure}

% \paragraph{The role of context}
\paragraph{Contextual dependencies of questions}

% To gain a better understanding of the contextual dependencies between questions in different positions in \Ours{}, we compute and visualize the token overlap percentage of the SQL answer (formal semantic representation of the question) between questions in different turns in Figure \ref{fig:sql_reuse}.
We visualize the percentage of token overlap between the SQL queries (formal semantic representation of the question) at different positions of a question sequence.
The heatmap shows that more information is shared between two questions that are closer to each other. This sharing increases among questions in later turns, where users tend to narrow down their questions to very specific needs.
This also indicates that resolving context references in our task is important.

Furthermore, the lighter color of the lower left 4 squares in the heatmap of Figure \ref{fig:sql_reuse} shows that most questions in an interaction have contextual dependencies that span within 3 turns. 
%\vic{How did we draw this conclusion?}. 
\citet{Reddy18} similarly report that the majority of context dependencies on the CoQA conversational question answering dataset are within 2 questions, beyond which coreferences from the current question are likely to be ambiguous with little inherited information. 
This suggests that % the length of 
3 turns on average are sufficient to capture most of the contextual dependencies between questions in \Ours{}.
%\vic{I'm worried that we cannot argue the length dependency on ATIS based on Figure 2, which is computed over the \Ours{} data.}
%\ty{but we also include the study of CoQA here to support our general conclusion.}
%\vic{I think the CoQA one is more like another fact. Claiming that ``3 turns is sufficient to capture most of the contextual dependencies between questions'' sounds too strong, if applied to all datasets. It is fine if the 3 threshold is obaserved from a plot based on the ATIS data.}
%\ty{but as we discussed ealier, all squares in the atis' heatmap would be very dark because the large overlap on db structures}

%Even though Table \ref{tb:data_stats} shows that \Ours{} has a shorter average interaction length (3) than ATIS (7), this 
%The length of 3 turns on average is sufficient to capture most of the contextual dependencies between questions.
% Moreover, the longer average interaction length of ATIS is  likely because no final end goal was enforced and the Wizard-of-OZ experiment was used to collect data.
% 2 turns for the CoQA conversational question answering dataset. 
%vic{It is necessary to make these comments about ATIS sounds objective and we should provide example for each of the claim make. We can insert a few examples here, no need to tabulated them or anything.}

\begin{figure}[!t]
    \vspace{-1.5mm}\hspace{-1mm}
    \centering
    \includegraphics[width=0.48\textwidth]{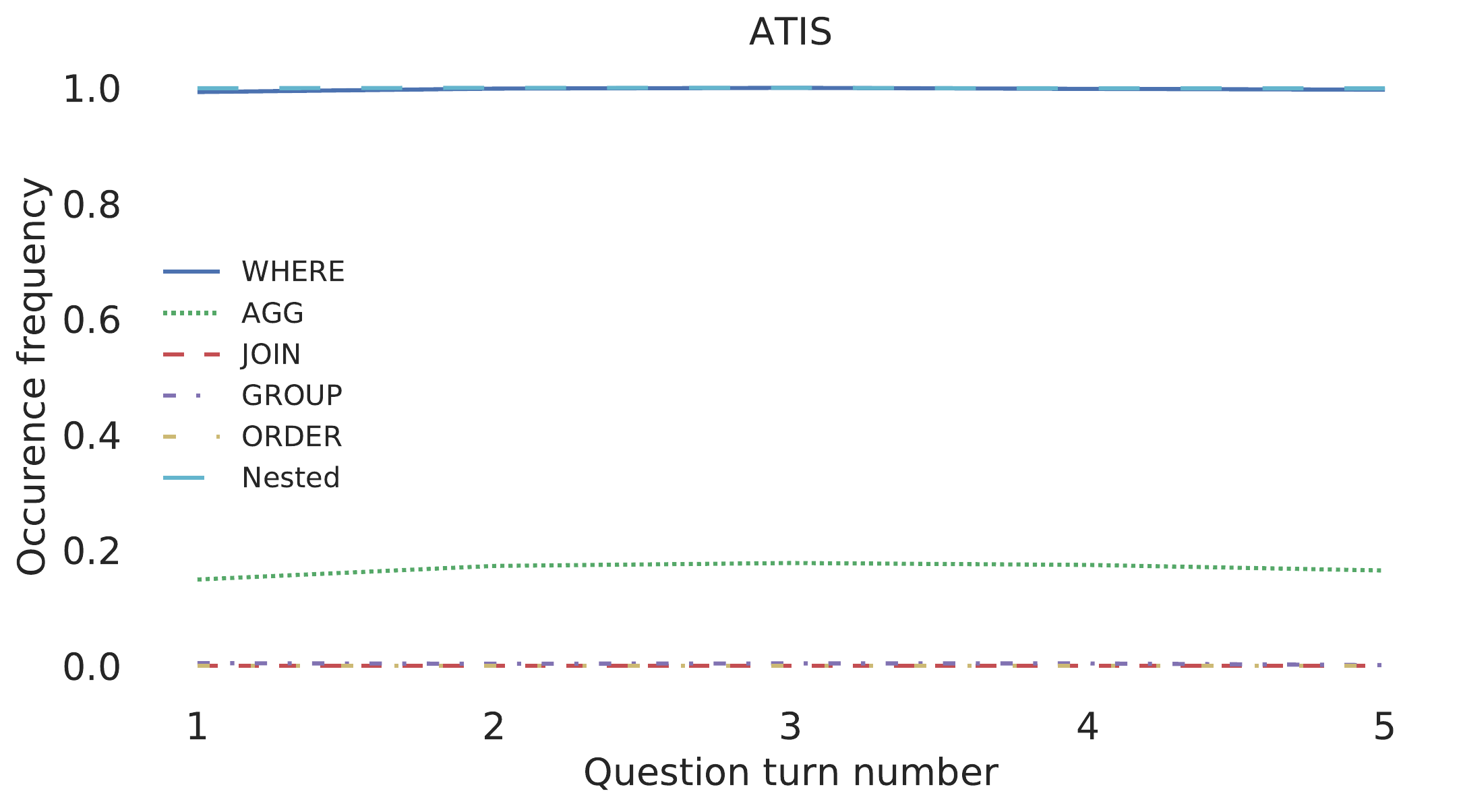}
    \includegraphics[width=0.48\textwidth]{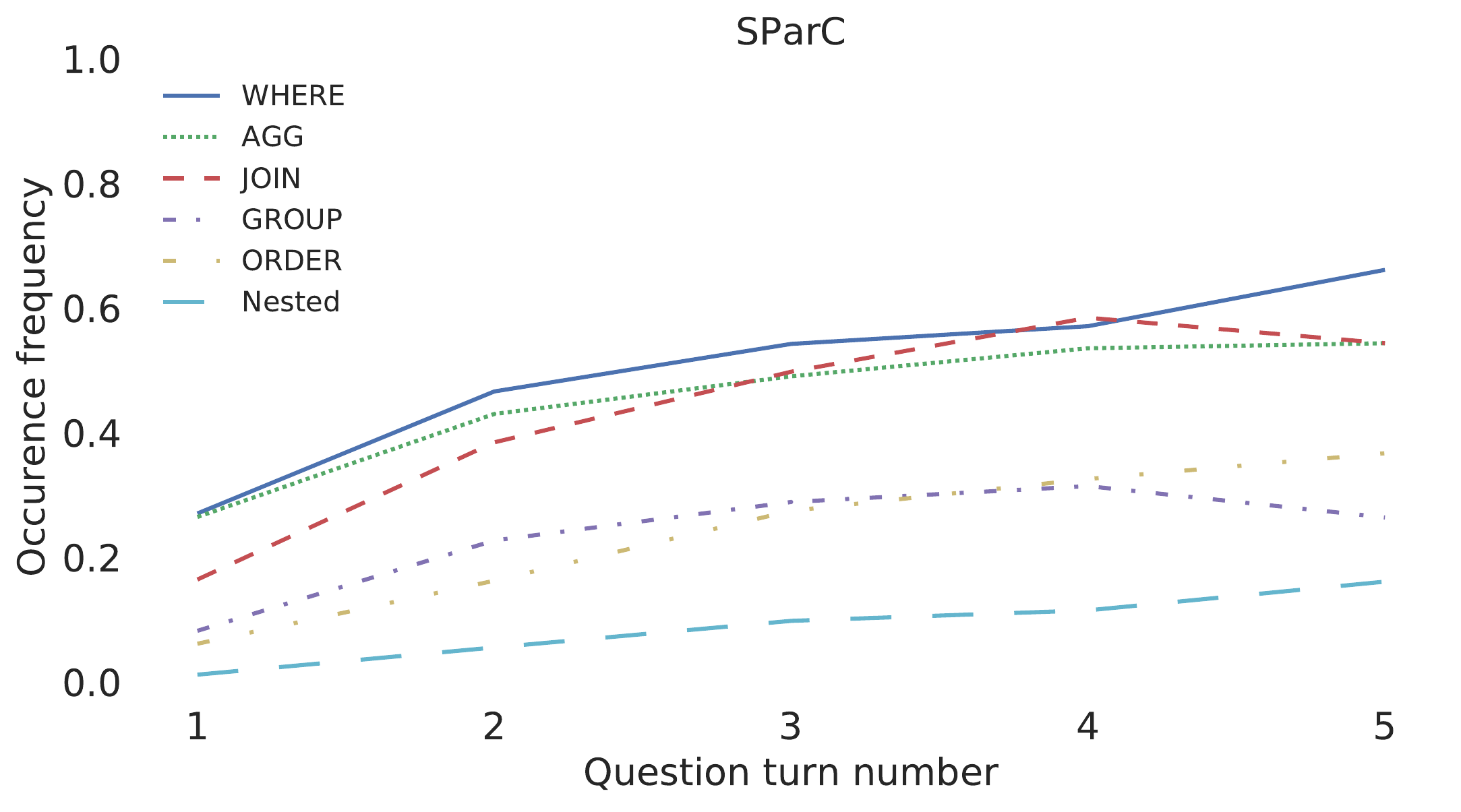}
    \caption{Percentage of question sequences that contain a particular SQL keyword at a given turn. The complexity of questions increases as interaction proceeds on \Ours{} as more SQL keywords are triggered. The same trend was not observed on ATIS.
    % \vic{Please follow the reviewer suggestion and mark the dataset name in each plot. This makes the plot easier to read comparing to only declaring them in the caption. You can put them in the top-right corner. And please draw the x- and y- axies using straight lines in both figures. Otherwise, the bottom lines in the top figure can be mistaken for the x-axis.}
    % The question position influences the semantic complexity of SQL answers in \Ours{} (lower) but not in ATIS (upper), suggesting that subsequent question turns in \Ours{}, but not ATIS, have changes in logical complexity between them. A SQL component is considered as occurrent if it appears at least once in the SQL answer
    }
    %More complex and specific questions are asked as the number of turns increases in our task, while questions in different turns in ATIS have similar semantic logic.\vic{Nice observation and very interesting characterization. However, need to explain what causes such differences.}
\label{fig:sql_turn}
\vspace{-1mm}
\end{figure}

We also plot the trend of common SQL keywords occurring in different question turns for both \Ours{} and ATIS (Figure~\ref{fig:sql_turn}) \footnote{Since the formatting used for the SQL queries are different in \Ours{} and ATIS, the actual percentages of \texttt{WHERE}, \texttt{JOIN} and \texttt{Nested} for ATIS are lower (e.g. the original version of ATIS may be highly nested, but queries could be reformatted to flatten them). Other SQL keywords are directly comparable.}. We show the percentage of question sequences that contain a particular SQL keyword at each turn. The upper figure in Figure \ref{fig:sql_turn} shows that the occurrences of all SQL keywords do not change much as the question turn increases in ATIS, which indicates that the SQL query logic % semantic contents of questions 
in different turns are very similar. We examined the data and find that most interactions in ATIS involve changes in the \texttt{WHERE} condition between question turns. This is likely caused by the fact that the questions in ATIS only involve flight booking, which typically triggers the use of the \textbf{\textit{refinement}} thematic relation. Example user utterances % questions 
from ATIS are \textit{``on american airlines"} or \textit{``which ones arrive at 7pm"} \cite{Suhr:18context}, which only involves changes to % or additions of 
the \texttt{WHERE} condition. 

In contrast, the lower figure demonstrates a clear trend that in \Ours{}, the occurrences of nearly all SQL components increase as question turn increases. This suggests that questions in subsequent turns tend to change the logical structures more significantly, which makes our task more interesting and challenging.%, such as \texttt{GROUP BY} and \texttt{ORDER BY}.

\paragraph{Contextual linguistic phenomena}
% \vic{Keep the original section heading here? It is different from the previous paragraph which analyzes SQL, this one analyzes natural language in context.}
We manually inspected 102 randomly chosen examples from our development set to study the % different 
thematic relations between questions.
Table \ref{tb:examples} shows the relation distribution.

We find that the most frequently occurring relation is \textbf{\textit{theme-entity}}, in which the current question focuses on % references 
the same entity (set) as the previous question but requests for some other property.
% In the first example of Figure \ref{fig:task}, the shared SQL tokens between $S_1$ and $S_2$ give the same result entities which are the dorms with a TV lounge.
Consider $Q_1$ and $Q_2$ of the first example shown in Figure~\ref{fig:task}. Their corresponding SQL representations ($S_1$ and $S_2$) have the same \texttt{FROM} and \texttt{WHERE} clauses, which harvest the same set of entities -- ``dorms with a TV lounge''.
But their \texttt{SELECT} clauses return different properties of the target entity set (number of the dorms in $S_1$ versus total capacity of the dorms in $S_2$).
$Q_3$ and $Q_4$ in this example also have the same relation.
The \textbf{\textit{refinement}} relation is also very common.
For example, $Q_2$ and $Q_3$ in the second example ask about the same entity set -- customers of a shipping company.
But $Q_3$ switches the search constraint from ``the most recent" in $Q_2$ to ``the first 5".

Fewer questions refer to previous questions by changing the entity (``Double Down" versus ``Keepers" in Table \ref{tb:examples}) but asking for the same property (\textbf{\textit{theme-property}}).
Even less frequently, some questions ask about % or are based on 
the answers of previous questions (\textbf{\textit{answer refinement/theme}}).
As in the last example of Table \ref{tb:examples}, the current question asks about the ``Statistics department'', which is one of the answers returned in the previous turn.
More examples with different thematic relations are provided in Figure \ref{fig:more_theme_examples} in the Appendix.

Interestingly, as the examples in Table \ref{tb:examples} have shown, many thematic relations are present without explicit linguistic markers.
This indicates that information tends to implicitly propagate through the interaction. % with implicit ways (without any coreference words).
Moreover, in some cases where the natural language question shares information with the previous question (e.g. $Q_2$ and $Q_3$ in the first example of Figure \ref{fig:task} form a \textbf{\textit{theme-entity}} relation), the corresponding SQL representations ($S_2$ and $S_3$) can be very different. One scenario in which this happens is when the property/constraint specification makes reference to additional entities described by separate tables in the database schema.

\paragraph{Semantic coverage} 
% As shown in Table \ref{tb:data_stats} \Ours{} is larger in terms of number of interaction sequences, questions and vocabulary size compared with ATIS. We note that the small vocabulary size of ATIS is likely due to its single domain and presence of many similar questions. 
As shown in Table \ref{tb:data_stats}, \Ours{} is larger in terms of number of unique SQL templates, % number of question sequences, 
vocabulary size and number of domains compared to ATIS. The smaller number of unique SQL templates and vocabulary size of ATIS is likely due to the domain constraint % its single domain 
and presence of many similar questions. 

\input{tables/data_sql.tex}
Table \ref{tab:data_sql} further compare the formal semantic representation in these two datasets in terms of SQL syntax component. % shows that \Ours{} has more diverse formal semantic coverage than ATIS as it contains a more complete set of SQL components.
While almost all questions in ATIS contain joins and nested subqueries, some commonly used SQL components are either absent (\texttt{ORDER BY}, \texttt{HAVING}, \texttt{SET}) or occur very rarely (\texttt{GROUP BY} and \texttt{AGG}). 
% This suggests that even though the ATIS dataset has longer average question length than \Ours{}, 
We examined the data and find that many questions in it has complicated syntactic structures mainly because the database schema requires joined tables and nested sub-queries, and the semantic diversity among the questions is in fact smaller.

%%%%%%%%%%% return here for future work %%%%%%%%%
%
% \vic{Future work experiment proposal -- compute the \# of different SQL templates in \Ours{} and ATIS.}
%\ty{add result comparison of questions in the same hardness level but in different poisitions.}
%\ty{cite some work like: https://arxiv.org/pdf/1704.05974.pdf and http://aclweb.org/anthology/D18-1192}
%}

% , and not due to the logical complexity of the questions. 
% In contrast, \Ours{} uses more varied SQL components, indicating  that \Ours{} has richer semantic coverage.

%Moreover, questions in ATIS are very ambiguous and require a lot prior knowledge on the database, which is overfitting to the answering behavior of a few annotators.
%This makes other SQL masters even harder to answer the questions in ATIS.
%To annotate ATIS, annotators must be very familiar with the flight database, otherwise others with strong SQL or database knowledge cannot answer most of the questions correctly.
%We agree that users might ask some ambiguous questions in a real-world database QA system, it is possible for the model to correctly answer these questions by overfitting to a specific domain.
%But that is impossible in cross-domain setting.
%The solution is to let users add some specific clarifications in late turns.

%In Table \ref{tab:data_sql}, we compare the semantic coverage of each dataset by counting SQL components in query labels. 

%Although SQL queries in ATIS are much longer in average, most of them repeatedly join multiple tables with simple logic structures.

\paragraph{Cross domain} 
% More importantly, as shown in Table \ref{tb:data_compare}, ATIS has databases only in the flight domain, which makes models built on it unable to generalize.
% As shown in Table \ref{tb:data_compare}, \Ours{} contains questions about 200 databases in 138 different domains including over 1,000 tables. In comparison, ATIS contains only 1 databases only in the flight domain, which makes models built on it unable to generalize.
As shown in Table \ref{tb:data_compare}, \Ours{} contains questions over 200 databases (1,020 tables) in 138 different domains.
In comparison, ATIS contains only one databases in the flight booking domain, which makes it unsuitable for developing models that generalize across domains.
Interactions querying different databases are shown in Figure \ref{fig:task} (also see more examples in Figure \ref{fig:more_examples} in the Appendix).
% difficult to train models that generalize across domains on top it.
% To make a model built on ATIS work for unseen domains, large amounts of data and model retraining is required, making this a very time-consuming and potentially unfeasible process.
% In contrast, \Ours{} contains questions about 200 databases in 138 different domains including over 1,000 tables. 
As in Spider, we split \Ours{} such that each database appears in only one of train, development and test sets.
Splitting by database requires the models to generalize not only to new SQL queries, but also to new databases and new domains.

\input{tables/data_split.tex}

% Therefore, \Ours{} enables the development of general-purpose database QA systems without additional SQL annotation required for an unseen database in a different domain.

%51% of the first questions in the sequences are column selections (e.g., “what are all of the teams?”). This number dwindles to just 18% when we look at the second question of each sequence, which indicates that the collected sequences start with general questions and progress to more specific ones.

%\input{tables/data_semantic.tex}

% we also observed different strategies among the subjects. the most common was to asking everything about an entity before turning to the next one, but some subjects preferred to ask about the value of a property for all the entities under discussion before turning to the next property.

%SQL represents some semantic meaning of the question.

%% file: tables/data_stats.tex
%\iffalse %%%%%%%%%%%
\begin{table}[t]
\centering
\scalebox{0.78}{
\begin{tabular}{ccc}
\Xhline{2\arrayrulewidth}
 & \textbf{\Ours{}} & ATIS \\\hline
Sequence \# & 4298 & 1658 \\
Question \# & 12,726 & 11,653 \\
Database \# & 200 & 1 \\
Table \# & 1020 & 27 \\
Avg. Q len & 8.1 & 10.2 \\
%SQL coverage \# & largest & medium \\
Vocab \# & 3794 & 1582 \\
Avg. turn \# & 3.0 & 7.0 \\
\Xhline{2\arrayrulewidth}
\end{tabular}}
\caption{Comparison of the statistics of context-dependent 
% semantic parsing 
text-to-SQL datasets.
%\vic{ty: too hard to define templates. Add a comparison of \# unique SQL templates in the two datasets, in order to support the ``Semantic coverage'' comparison in section 4.}
} 
\label{tb:data_stats}
\end{table}
%\fi %%%%%%%%%%%

%extension our dataset is diffferent pos dataset, is more application
%\vic{For a dataset paper this table describes how much contribution the new dataset has added over previous data hence is very, very important. Currently the table present mixed results. Issue 1: Comparisng w/ AITS, \Ours{} is bigger in terms of \# questions, \# domains and vocabulary size, but not order of magnitudes larger. On the other hand the \# of turns are smaller in \Ours{}, hence need to carefully explain why shorter question sequences are obtained (My guess is that ATIS used the WIzard-of-Oz method and does not have the end goal restriction as \Ours{}.). Issue 2: Comparing with \SQA{}, \Ours{} is smaller in terms of \# questions and vocabulary size (significantly smaller). But \SQA{} is not a text-to-SQL dataset and it is hard to interpret what it means by 0/982 domains, hence it is better to remove \SQA{} from the comparison here. I hence changed the table title from semantic parsing datasets to text-to-SQL datasets.}

%% file: tables/data_sql.tex
\begin{table}[t]
\centering
\small
\begin{tabular}{cccc}
\Xhline{2\arrayrulewidth}
SQL components & \Ours{} & ATIS \\\hline
\# \texttt{WHERE} & 42.8\% & 99.7\% \\
\# \texttt{AGG} & 39.8\% & 16.6\% \\
\# \texttt{GROUP} & 20.1\% & 0.3\%\\
\# \texttt{ORDER} & 17.0\% & 0.0\% \\
\# \texttt{HAVING} & 4.7\% & 0.0\%\\
\# \texttt{SET} & 3.5\% & 0.0\% \\
\# \texttt{JOIN} & 35.5\% & 99.9\% \\
\# \texttt{Nested} & 5.7\% & 99.9\% \\
%\# \texttt{Others} & 23.0 & 99.9\\
\Xhline{2\arrayrulewidth}
\end{tabular}
\caption{Distribution of SQL components in SQL queries. SQL queries in \Ours{} cover all SQL components, whereas some important SQL components like \texttt{ORDER} are missing from ATIS.}
\label{tab:data_sql}
\end{table}

%% file: tables/data_split.tex
\begin{table}[h!]
\centering
\small
\begin{tabular}{cccc}
\Xhline{2\arrayrulewidth}
 & Train & Dev & Test \\\hline
\# Q sequences & 3034 & 422 & 842\\
\# Q-SQL pairs & 9025 & 1203 & 2498 \\
\# Databases & 140 & 20 & 40 \\
\Xhline{2\arrayrulewidth}
\end{tabular}
\caption{Dataset Split Statistics}
\label{tab:data_split}
\end{table}

%% file: methods.tex
\section{Methods}
\label{sec:baselines}
% \vic{Would it make sense to draw the NN architectures and add the figures to the appendix?}
% In order to analyze the difficulty and demonstrate the purpose of our corpus, we experiment with two state-of-the-art semantic parsing models extended to our task.
% To benchmark the difficulty of our dataset, we experiment with 
We extend two state-of-the-art semantic parsing models to the cross-domain, context-dependent setup of \Ours{} and benchmark their performance. At each interaction turn $i$, given the current question $\Bar{x}_i=\langle x_{i,1},\hdots,x_{i,|\Bar{x}_i|}\rangle$, the previously asked questions $\Bar{I}[:i-1] = \{\Bar{x}_1,\hdots,\Bar{x}_{i-1}\}$ and the database schema $C$, the model generates the SQL query $\Bar{y}_i$.
%: (1) \seqcon{}, a domain-specific context-dependent model for ATIS proposed by \newcite{Suhr:18context}, and (2) SyntaxSQLNet introduced by \cite{Yu&al.18.emnlp.syntax} for cross-domain context-independent model on Spider.\vic{Should this be CD-SyntaxSQLNet since it is extended to handle context-dependent generation?}
%As our task differs from both ATIS \cite{Hemphill90,Dahl94} in terms of cross-domain and semantic richness and Spider \cite{Yu18emnlp} in terms of context-dependent, we adapted these models to our task by enabling \seqcon{} to handle unseen domains and SyntaxSQLNet to consider previous questions and SQL history.
%as follows.
% \vic{Add a sentence summarizing the major takeaways after observing the baseline performance.}

\subsection{Seq2Seq with turn-level history encoder (\seqcon{})} 

This is a cross-domain Seq2Seq based text-to-SQL model extended with the turn-level history encoder proposed in~\citet{Suhr:18context}.

\paragraph{Turn-level history encoder} Following~\citet{Suhr:18context}, at turn $i$, we encode each user question $\Bar{x}_t\in\Bar{I}[:t-1]\cup\{\Bar{x}_i\}$ using an utterance-level bi-LSTM, $\text{LSTM}^{E}$. The final hidden state of $\text{LSTM}^{E}$, $\newvec{h}^E_{t,|\Bar{x}_t|}$, is used as the input to the turn-level encoder, $\text{LSTM}^{I}$, a uni-directional LSTM, to generate the discourse state $\newvec{h}_t^I$. The input to $\text{LSTM}^{E}$ at turn $t$ is the question word embedding concatenated with the discourse state at turn $t-1$ ($[\newvec{x}_{t,j}, \newvec{h}_{t-1}^I]$), which enables the flow of contextual information.

\paragraph{Database schema encoding} For each column header in the database schema, we concatenate its corresponding table name and column name separated by a special dot token (i.e., \texttt{table\_name.column\_name}), and use the average word embedding\footnote{We use the 300-dimensional GloVe \cite{pennington14} pretrained word embeddings.} of tokens in this sequence as the column header embedding $\newvec{h}^{C}$.

\paragraph{Decoder} The decoder is implemented with another LSTM ($\text{LSTM}^D$) with attention to the $\text{LSTM}^{E}$ representations of the questions in $\eta$ previous turns ($\eta$ is a hyperparameter). At each decoding step, the decoder chooses to generate either a SQL keyword (e.g., \texttt{select}, \texttt{where}, \texttt{group by}) or a column header.
To achieve this, we use separate layers to score SQL keywords and column headers, and finally use the softmax operation to generate the output probability distribution over both categories.

\subsection{SyntaxSQLNet with history input (\syncon{})}
SyntaxSQLNet is a syntax tree based neural model for the complex and cross-domain context-independent text-to-SQL task introduced by \citet{Yu&al.18.emnlp.syntax}. % for Spider. 
The model consists of a table-aware column attention encoder and a SQL-specific syntax tree-based decoder. The decoder adopts a set of inter-connected neural modules to generate different SQL syntax components. % with SQL generation path history. 
%Since it was originally proposed for the Spider task, 
% Since it only handles single-turn questions, we provide the model with a short interaction history so as to consider context dependencies.
% We encode a history of the natural language utterance and the associated SQL query response (gold at training time, predicted at test time) for that utterance in each module (\syncon{}). 
% For the model to leverage context information, we 

We extend this model by providing the decoder with the encoding of the previous question ($\Bar{x}_{i-1}$) as additional contextual information.
%and the associated SQL query response (gold at training time, predicted at test time) \vic{How is the history ``provided''? Is it simply concatenated to the the original encoder input?} 
% The encoding method for both is identical to the encoding used for the current utterance (ie. a bidirectional LSTM for utterance).
Both $\Bar{x}_i$ and $\Bar{x}_{i-1}$ are encoded using bi-LSTMs (of different parameters) with the column attention mechanism proposed by~\citet{Yu&al.18.emnlp.syntax}.
%\vic{Do the biLSTMs for previous utterance, previous SQL and current utterance share the same parameters?}
% We also employ the column attention mechanism to compute final representations of the previous question weighted by attentions with columns.
%\vic{What is column attention? For this attention, what is the query, what is the key? Most importantly, what is the output of this attention? Do the representations of previous question, previously SQL and current question all gets updated by this attention?}
% After that, we follow the same steps as the current question to incorporate these representations in each module.
We use the same math formulation to inject the representations of $\Bar{x}_i$ and $\Bar{x}_{i-1}$ to each syntax module of the decoder.
% The encoded input thus provides the model, at decoding time, access not only to the utterance current question and the database schema, but also to the previous context.
% The tree decoder is also used to recursively generate SQL tokens.
% Data augmentation method is not used.
% Also, experimental settings are the same as \cite{Yu&al.18.emnlp.syntax} used.

%\vic{How to provide the encoded history to the model? Attention?}
%\vic{Reviewer will almost certainly raise the question that why we only evaluate SyntaxSQLNet w/ one-step history. This makes the model comparisons in section 6 more difficult since the model variations are not controlled. For example, give that \syncon{} achieves lower joint goal accuracy, it is very likely due to the fact that \syncon{} only looks back 1-step in history. However, other hidden factors could also cause this s.a. the way how the SyntaxSQLNet decoder works makes it more difficult to absorb contextual information. Need to be very cautious when describing the results s.t. the reader don't feel we mixed up the result analysis.}
%\vic{How was both models trained? Random initialization? BERT?}

More details of each baseline model can be found in the Appendix. And we opensource their implementations for reproducibility.

%% file: results.tex
\section{Experiments}
\label{sec:results}

\subsection{Evaluation Metrics}
\label{sec:eval}
Following \citet{Yu18emnlp}, we use the exact set match metric to compute the accuracy between gold and predicted SQL answers. 
% To avoid ordering issues \cite{Xu2017}, 
Instead of simply employing string match, \citet{Yu18emnlp} decompose predicted queries into different SQL clauses such as \texttt{SELECT}, \texttt{WHERE}, \texttt{GROUP BY}, and \texttt{ORDER BY} and compute scores for each clause using set matching separately\footnote{Details of the evaluation metrics can be found at \scriptsize \url{https://github.com/taoyds/spider/tree/master/evaluation_examples}}.
We report the following two metrics: \textit{question match}, the exact set matching score over all questions, and \textit{interaction match}, the exact set matching score over all interactions. The exact set matching score is 1 for each question only if all predicted SQL clauses are correct, and 1 for each interaction only if there is an exact set match for every question in the interaction.

%To better understand model performance on our task, we also report the results of the following evaluation metrics: 
%\paragraph{Turn request}
%It computes the overall exact set match accuracy on all the questions in all question sequences without considering positions.
%\paragraph{Joint goal}
%It computes the overall exact set match accuracy on all question sequences. 
%The sequence prediction is right only when all questions are predicted correctly.
%This shows the ability of how the system successfully accomplishes users' final query goals.

\subsection{Results}

\input{tables/result_main.tex}

We report the overall results of \seqcon{} and SyntaxSQLNet on the development and the test data in Table \ref{tb:res_main}.
The context-aware models (\seqcon{} and \syncon{}) significantly outperforms the context-agnostic 
SyntaxSQLNet (\syninp{}). % by considering context.
The last two rows form a controlled ablation study, where without accessing to the previous question, the test set performance of SyntaxSQLNet decreases from 20.2\% to 16.9\% on \textit{question match} and from 5.2\% to 1.1\% on \textit{interaction match}, which indicates that context is a crucial aspect of the problem.

We note that \syncon{} scores higher in \textit{question match} but lower in \textit{interaction match} compared to \seqcon{}.
% A reason could be that \syncon{} predicts more questions in the early turns of an interaction  correctly because it uses a column attention mechanism and cross-domain encoders \cite{Yu&al.18.emnlp.syntax}, 
A closer examination shows that \syncon{} predicts more questions correctly in the early turns of an interaction (Table~\ref{tab:res_turn}), which results in its overall higher \textit{question match} accuracy. A possible reason for this is that \syncon{} adopts a stronger context-agnostic text-to-SQL module (SyntaxSQLNet vs. Seq2Seq adopted by \seqcon{}).
The higher performance of  \seqcon{} on \textit{interaction match} can be attributed to better incorporation of information flow between questions by using turn-level encoders~\cite{Suhr:18context}, which is possible to encode the history of all previous questions comparing to only single one previous question in \syncon{}.
%\vic{Add a sentence summarizing the exact benefit of turn-level encoders.}
Overall, the lower performance of the two extended context-dependent models shows the difficulty of \Ours{} and that there is ample room for improvement. % for the task of context-dependent semantic parsing.
%Especially, less than 10\% accuracy on answering all related questions from a user, which limits the real-world application of natural language interfaces to databases. 
%51% of the first questions in the sequences are column selections (e.g., “what are all of the teams?”). This number dwindles to just 18% when we look at the second question of each sequence, which indicates that the collected sequences start with general questions and progress to more specific ones.

\input{tables/result_turns.tex}

\paragraph{Performance stratified by question position}
To gain more insight into how question position affects the performance of the two models, we report their performances on questions in different positions in Table \ref{tab:res_turn}. 
Questions in later turns of an interaction in general have % greater informational demands due to more prior questions 
greater dependency over previous questions and also greater risk for error propagation.
The results show that both \seqcon{} and \syncon{} consistently perform worse as the question turn increases, suggesting that both models struggle to deal with information flow from previous questions and accumulate errors. % from previous predictions.
%again indicates that our dataset exposes a challenge on handling contextual dependencies among questions in the sequence.
Moreover, \syncon{} significantly outperforms \seqcon{} on questions in the first turn,
but the advantage disappears in later turns (starting from the second turn), % right after the first questions 
% because \syncon{} is less able to handle information flows from prior questions and predictions, which is expected since \seqcon{} was originally developed on ATIS (context-aware), whereas \syncon{} was developed on Spider (context-agnostic).
which is expected because the context encoding mechanism of \syncon{} is less powerful than the turn-level encoders adopted by \seqcon{}.

\input{tables/result_hardness.tex}
\paragraph{Performance stratified by SQL difficulty}
% \citet{Yu18emnlp} % automatically 
% groups questions in Spider into different difficulty levels based on the complexity of their corresponding SQL representations. 
We group individual questions in \Ours{} into different difficulty levels based on the complexity of their corresponding SQL representations using the criteria proposed in \citet{Yu18emnlp}.
As shown in Figure 3, the questions turned to get harder as interaction proceeds, more questions with hard and extra hard difficulties appear in late turns. 
%\vic{Is there a correlation between difficulty and turns? Does the questions turned to get harder as interaction proceeds? Figure 3 shows this in some extent but not exactly. Can we add a plot in appendix to show this?}
Table \ref{tab:res_hard} shows the performance of the two models across each difficulty level. As we expect, the models perform better when the user request is easy.
Both models fail on most hard and extra hard questions. % requests, which appear in the last turn.
Considering that the size and question types of \Ours{} are very close to Spider, the relatively lower performances of \syncon{} on medium, hard and extra hard questions in Table~\ref{tab:res_hard} comparing to its performances on Spider (17.6\%, 16.3\%, and 4.9\% respectively) indicates that \Ours{} introduces additional challenge by introducing context dependencies, which is absent from Spider.

\input{tables/result_theme.tex}
\paragraph{Performance stratified by thematic relation}
Finally, we report the model performances across thematic relations computed over the 102 examples summarized in Table \ref{tb:examples}.
The results (Table \ref{tab:res_thme}) show that the models, in particular \syncon{}, perform the best on the \textbf{\textit{answer refinement/theme}} relation.
% A possible reason is that these questions rely less on context dependencies than questions of other thematic relations. For example, the Cur\_Q for \textit{theme/refinement-answer} in Table \ref{tb:examples} is asking a property of a subset of the entity in the answer of the previous question, and shares little co-referred information with the Prev\_Q. 
A possible reason for this is that questions in the \textbf{\textit{answer theme}} category can often be interpreted without reference to previous questions since the user tends to state the theme entity explicitly. Consider the example in the bottom row of Table \ref{tb:examples}. The user explicitly said ``Statistics department'' in their question, which belongs to the answer set of the previous question \footnote{As pointed out by one of the anonymous reviewers, there are less than 10 examples of \textit{answer refinement/theme} relation in the 102 analyzed examples. We need to see more examples before concluding that this phenomenon is general.}.
The overall low performance for all thematic relations (\textbf{\textit{refinement}} and \textbf{\textit{theme-property}} in particular) indicates that the two models still struggle on properly interpreting the question history.
% On the other hand, the two models struggle on questions that have \textit{refinement} or \textit{theme-property} relations with previous questions. 
% This is likely due to models needing to refer back in the question history to answer the current question.% \vic{Give an example illustrating what ``further back'' means.}

%% file: tables/result_main.tex
\begin{table}[h!]
\centering
\resizebox{\columnwidth}{!}{
\begin{tabular}{ccccc}
\Xhline{2\arrayrulewidth}
Model & \multicolumn{2}{c}{Question Match} & \multicolumn{2}{c}{Interaction Match} \\
              & Dev & Test &  Dev & Test \\ \hline
\seqcon{}          & 17.1 & 18.3 & \bf{6.7} & \bf{6.4} \\
\syncon{} & \bf{18.5} & \bf{20.2} & 4.3 & 5.2 \\
\syninp{} & 15.2 & 16.9 & 0.7 & 1.1 \\
\Xhline{2\arrayrulewidth}
\end{tabular}
}
\caption{Performance of various methods over all questions (\textit{question match}) and all interactions (\textit{interaction match}).} 
\label{tb:res_main}
\end{table}

%% file: tables/result_turns.tex
\begin{table}[ht!]
\centering
\small
\begin{tabular}{ccc}
\Xhline{2\arrayrulewidth}
Turn \# & \seqcon{} & \syncon{} \\\hline
1 (422) & 31.4 & 38.6 \\ %35.1
2 (422) & 12.1 & 11.6 \\ %5.2
3 (270) & 7.8 & 3.7 \\ %4.8
$\ge 4$ (89) & 2.2  & 1.1 \\ %0
\Xhline{2\arrayrulewidth}
\end{tabular}
\caption{Performance stratified by question turns on the development set. The performance of the two models decrease as the interaction continues.}
\label{tab:res_turn}
\end{table}
%check &SyntaxSQLNet-inp

%% file: tables/result_hardness.tex
\begin{table}[h!]
\centering
\small
\begin{tabular}{ccc}
\Xhline{2\arrayrulewidth}
Goal Difficulty & \seqcon{} & \syncon{}\\\hline
Easy (483) & 35.1 & \bf{38.9} \\
Medium (441) & 7.0 & \bf{7.3} \\
Hard (145) & \bf{2.8} & 1.4 \\
Extra hard (134) & \bf{0.8} & 0.7 \\
\Xhline{2\arrayrulewidth}
\end{tabular}
\caption{Performance stratified by question difficulty on the development set. The performances  of the two models decrease as questions are more difficult.}
\label{tab:res_hard}
\end{table}

%% file: tables/result_theme.tex
\begin{table}[ht!]
\centering
\small
\begin{tabular}{ccc}
\Xhline{2\arrayrulewidth}
Thematic relation & \seqcon{} & \syncon{}\\\hline
Refinement & 8.4 & 6.5 \\
Theme-entity & 13.5 & 10.2 \\
Theme-property & 9.0 & 7.8 \\
answer refine./them.& 12.3 & 20.4 \\
\Xhline{2\arrayrulewidth}
\end{tabular}
\caption{Performance stratified by thematic relations. The models perform best on the \textit{answer refinement/theme} relation, but do poorly on the \textit{refinement} and \textit{theme-property} relations.}
\label{tab:res_thme}
\end{table}

%% file: conclusion.tex
\section{Conclusion}
In this paper, we introduced \Ours{}, a large-scale dataset of % conversational interactions 
context-dependent questions over a number of databases in different domains annotated with the corresponding SQL representation.
The dataset features wide semantic coverage and a diverse set of contextual dependencies between questions. % in the same sequence, 
It also introduces unique challenge in mapping context-dependent questions to SQL queries in unseen domains. 
% The associated task introduces a challenge in mapping context-dependent questions to SQL queries in unseen domains. 
We experimented with two competitive context-dependent semantic parsing approaches on \Ours{}. The model accuracy is far from satisfactory and stratifying the performance by question position shows that both models degenerate in later turns of interaction, suggesting the importance of better context modeling.  % and observed relatively weak performance, which suggests ample challenges for future research.
% This, together with our detailed data analysis, exemplifies the complexity of the data. 
The dataset, baseline implementations and leaderboard are publicly available at \url{https://yale-lily.github.io/sparc}.

%% file: acknowledgements.tex
\section*{Acknowledgements}
We thank Tianze Shi for his valuable discussions and feedback.
We thank the anonymous reviewers for their thoughtful detailed comments.

%% file: appendix.tex
\section{Appendices}
\label{sec:appendix}

\subsection{Additional Baseline Model Details}
\paragraph{CD-Seq2Seq}
% \ty{Rui adds more details here}
We use a bi-LSTM, $\text{LSTM}^{E}$, to encode the user utterance at each turn.
At each step $j$ of the utterance, $\text{LSTM}^{E}$ takes as input the word embedding and the discourse state $\newvec{h}^{I}_{i-1}$ updated for the previous turn $i-1$:
\begin{equation*}
    \newvec{h}^{E}_{i,j} = \text{LSTM}^{E}([\newvec{x}_{i,j}; \newvec{h}^{I}_{i-1}], \newvec{h}^{E}_{i,j-1})
\end{equation*}
where $i$ is the index of the turn and $j$ is the index of the utterance token.
The final hidden state $\text{LSTM}^{E}$ is used as the input of a uni-directional LSTM, $\text{LSTM}^{I}$, which is the interaction level encoder:
\begin{equation*}
    \newvec{h}^{I}_{i} = \text{LSTM}^{I}(\newvec{h}^{E}_{|x_i|}, \newvec{h}^{I}_{i-1}).
\end{equation*}

For each column header, we concatenate its table name and its column name separated by a special dot token (i.e., \texttt{table\_name.column\_name}), and the column header embedding $\newvec{h}^{C}$ is the average embeddings of the words.

The decoder is implemented as another LSTM with hidden state $\newvec{h}^{D}$.
We use the dot-product based attention mechanism to compute the context vector.
At each decoding step $k$, we compute attention scores for all tokens in $\eta$ previous turns (we use $\eta=5$) and normalize them using softmax.
Suppose the current turn is $i$, and consider the turns of $0,\dots,\eta-1$ distance from turn $i$.
We use a learned position embedding $\phi^{I}(i-t)$ when computing the attention scores.
The context vector is the % attention 
weighted sum of the concatenation of the token embedding and the position embedding:
\begin{align*}
\begin{split}
    s_{k}(t,j) & = [\newvec{h}^{E}_{t,j};\phi^{I}(i-t)]\newvec{W}_{\text{att}}\newvec{h}^{D}_{k} \\
    \alpha_{k} & = \softmax(s_{k}) \\
    \newvec{c}_{k} & = \sum_{t=i-h}^{i}\sum_{j=1}^{|x_t|}\alpha_{k}(t,j)[\newvec{h}^{E}_{t,j};\phi^{I}(i-t)]\\
\end{split}
\end{align*}

% In the output layer, our decoder 
At each decoding step, the sequential decoder chooses to generate a SQL keyword (e.g., \texttt{select}, \texttt{where}, \texttt{group by}, \texttt{order by}) or a column header.
To achieve this, we use separate layers to score SQL keywords and column headers, and finally use the softmax operation to generate the output probability distribution:
\begin{align*}
\begin{split}
\newvec{o}_{k}             & = \tanh([\newvec{h}^{D}_{k};\newvec{c}_k]\newvec{W}_{o}) \\
\newvec{m}^{\text{SQL}}    & = \newvec{o}_{k}\newvec{W}_{\text{SQL}} + \newvec{b}_{\text{SQL}} \\
\newvec{m}^{\text{column}} & = \newvec{o}_{k}\newvec{W}_{\text{column}}\newvec{h}^{C} \\
P(y_k)                     & = \softmax([\newvec{m}^{\text{SQL}}; \newvec{m}^{\text{column}}]) \\
\end{split}
\end{align*}

It's worth mentioning that we experimented with a SQL segment copying model similar to the one proposed in~\citet{Suhr:18context}. We implement our own segment extraction procedure by extracting \texttt{SELECT}, \texttt{FROM}, \texttt{GROUP BY}, \texttt{ORDER BY} clauses as well as different conditions in \texttt{WHERE} clauses.
In this way, we can extract 3.9 segments per SQL on average.
However, we found that adding segment copying does not significantly improve the performance because of error propagation.
Better leveraging previously generated SQL queries remains an interesting future direction for this task.
% We do not use anonymization scoring from \newcite{Suhr:18context} in because our data set and task employs anonoymized database values.

\paragraph{SyntaxSQL-con}
 As in \cite{Yu&al.18.emnlp.syntax}, the following is defined to compute the conditional embedding $\mathbf{H}_{1/2}$ of an embedding $\mathbf{H}_1$ given another embedding $\mathbf{H}_2$:
$$
\mathbf{H}_{1/2} = \softmax(\mathbf{H}_{1}\mathbf{W} \mathbf{H}_{2}^\top) \mathbf{H}_{1}.
$$
Here $\mathbf{W}$ is a trainable parameter.
In addition, a probability distribution from a given score matrix $\mathbf{U}$ is computed by
$$
\mathcal{P}(\mathbf{U}) = \softmax \left(\mathbf{V} \textbf{tanh}(\mathbf{U}) \right),
$$
where $\mathbf{V}$ is a trainable parameter. 
To incorporate the context history, we encode the question right before the current question and add it to each module as an input.
% For example, if we denote the hidden states of LSTM on embeddings of the previous one question as $\mathbf{H}_{\textrm{PQ}}$, the number of columns and which ones to select in the COL module are computed by
For example, the COL module of SyntaxSQLNet is extended as following. $\mathbf{H}_{\textrm{PQ}}$ denotes the hidden states of LSTM on embeddings of the previous one question and the $\mathbf{W}^{\textrm{num}}_{3} {\mathbf{H}^{\textrm{num}}_{\textrm{PQ/COL}}}^\top$ and $\mathbf{W}^{\textrm{val}}_{4} {\mathbf{H}^{\textrm{val}}_{\textrm{PQ/COL}}}^\top$ terms add history information to prediction of the column number and column value respectively.
\begin{gather*}
\resizebox{\hsize}{!}{$
P^{\textrm{num}}_{\textrm{COL}} = \mathcal{P} \left(\mathbf{W}^{\textrm{num}}_{1} {\mathbf{H}^{\textrm{num}}_{\textrm{Q/COL}}}^\top + {\mathbf{W}^{\textrm{num}}_{2} \mathbf{H}^{\textrm{num}}_{\textrm{HS/COL}}}^\top
+ \mathbf{W}^{\textrm{num}}_{3} {\mathbf{H}^{\textrm{num}}_{\textrm{PQ/COL}}}^\top\right)$}
\\
\resizebox{\hsize}{!}{$
P^{\textrm{val}}_{\textrm{COL}} = \mathcal{P} \left( \mathbf{W}^{\textrm{val}}_{1} {\mathbf{H}^{\textrm{val}}_{\textrm{Q/COL}}}^\top + \mathbf{W}^{\textrm{val}}_{2} {\mathbf{H}^{\textrm{val}}_{\textrm{HS/COL}}}^\top + \mathbf{W}^{\textrm{val}}_{3} {\mathbf{H}_{\textrm{COL}}}^\top
+ \mathbf{W}^{\textrm{val}}_{4} {\mathbf{H}^{\textrm{val}}_{\textrm{PQ/COL}}}^\top
\right)$}
\end{gather*}

\subsection{Additional Data Examples}
\begin{figure*}[t!]
    % \vspace{-1.5mm}\hspace{-1mm}
    \centering
    \includegraphics[width=0.85\textwidth]{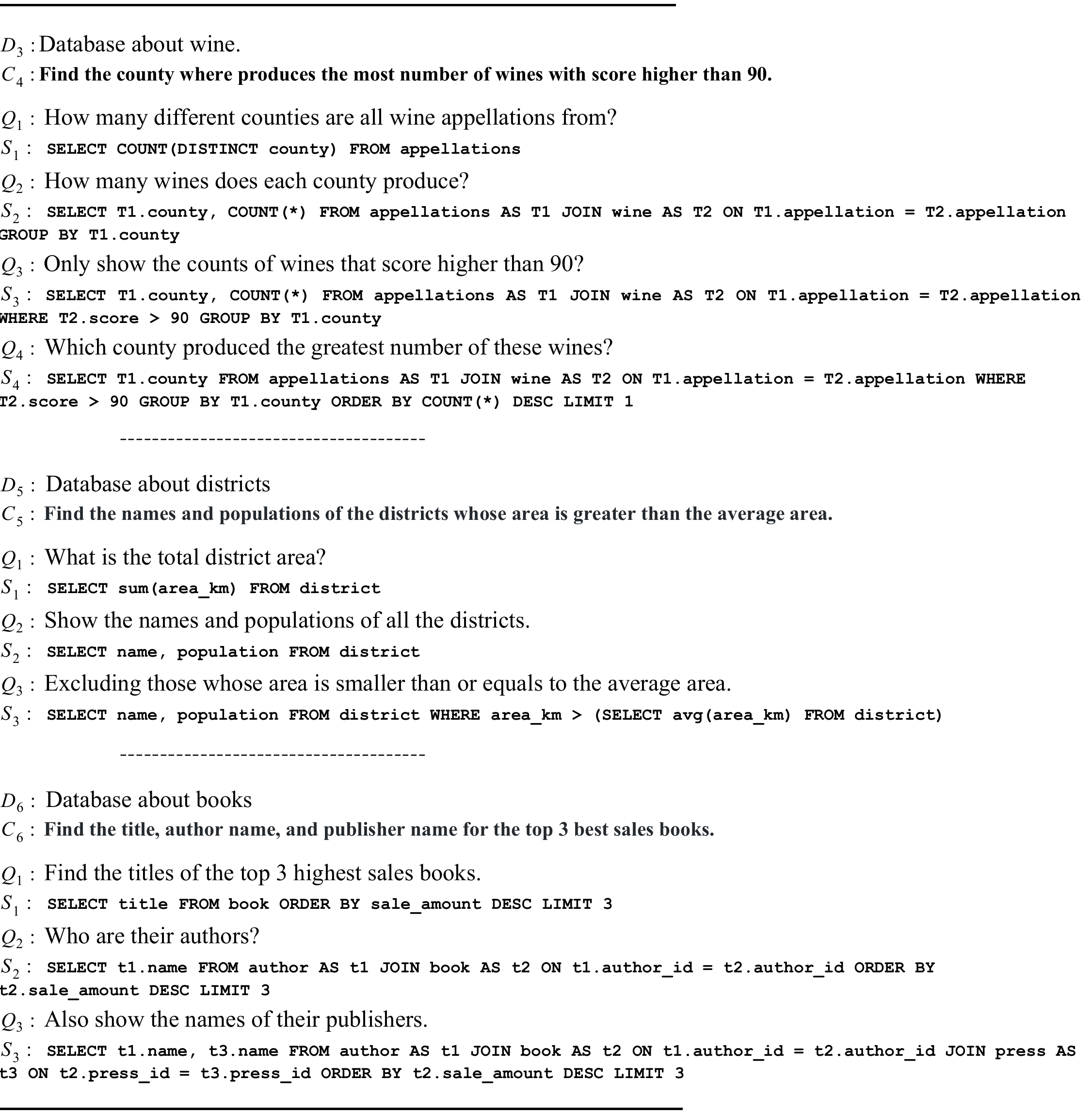}
    \vspace{-2mm}
    \caption{More examples in \Ours{}.}
\label{fig:more_examples}
\vspace{-2.54mm}
\end{figure*}

\begin{figure*}[t!]
    \vspace{6mm}% \hspace{-1mm}
    \centering
    \includegraphics[width=0.85\textwidth]{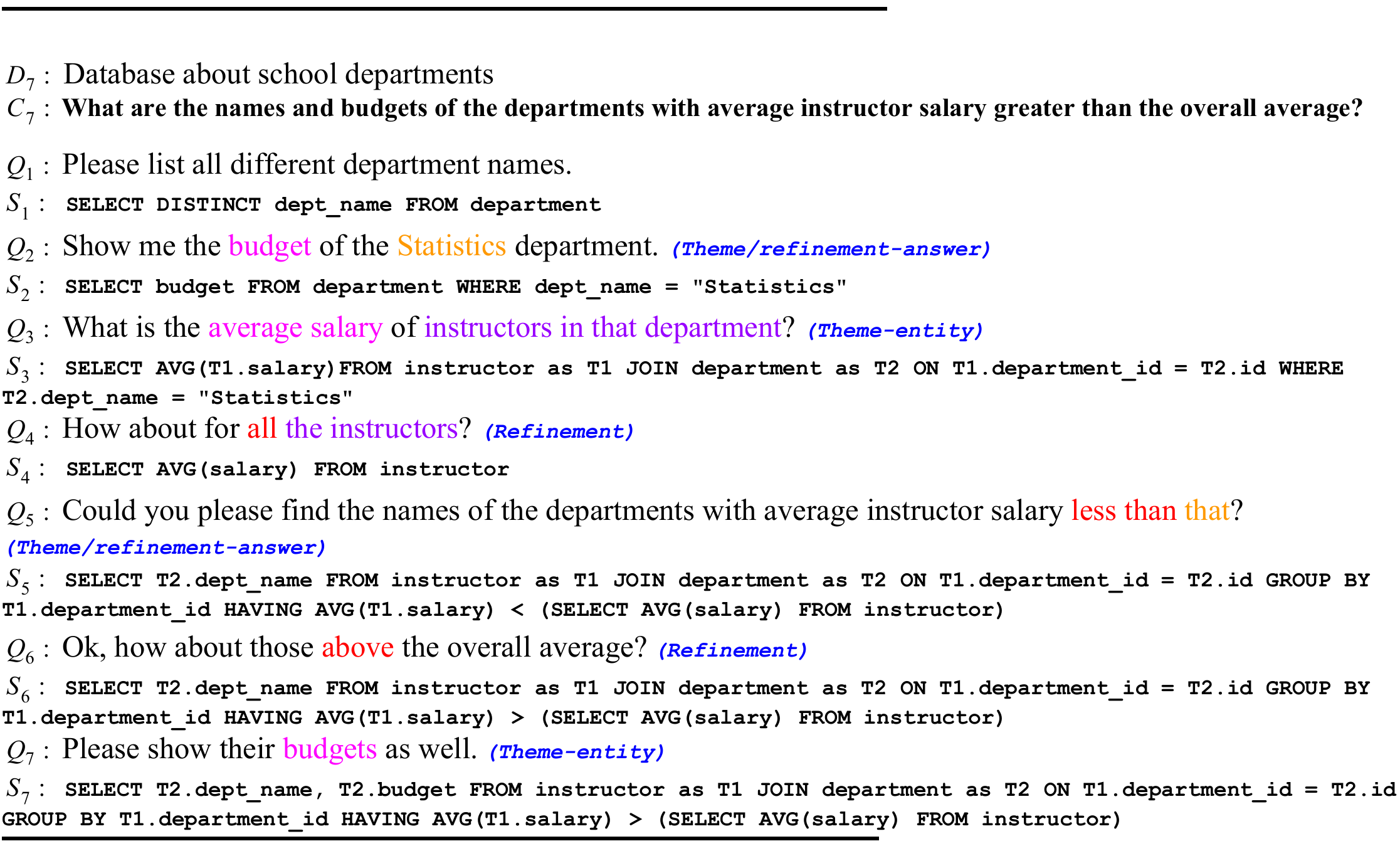}
    \vspace{-2mm}
    \caption{Additional example in \Ours{} annotated with different thematic relations. Entities (\textcolor{purple}{purple}), properties (\textcolor{magenta}{magenta}), constraints (\textcolor{red}{red}), and answers (\textcolor{orange}{orange}) are colored.}
\label{fig:more_theme_examples}
\vspace{-2.54mm}
\end{figure*}

We provide additional \Ours{} examples in Figure \ref{fig:more_examples} and examples with different thematic relations in Figure \ref{fig:more_theme_examples}.

%% file: acl2019.bbl
\begin{thebibliography}{38}
\expandafter\ifx\csname natexlab\endcsname\relax\def\natexlab#1{#1}\fi

\bibitem[{Artzi and Zettlemoyer(2013)}]{artzi13}
Yoav Artzi and Luke Zettlemoyer. 2013.
\newblock Weakly supervised learning of semantic parsers for mapping
  instructions to actions.
\newblock \emph{Transactions of the Association forComputational Linguistics}.

\bibitem[{Berant and Liang(2014)}]{Berant14}
Jonathan Berant and Percy Liang. 2014.
\newblock Semantic parsing via paraphrasing.
\newblock In \emph{Proceedings of the 52nd Annual Meeting of the Association
  for Computational Linguistics (Volume 1: Long Papers)}, pages 1415--1425,
  Baltimore, Maryland. Association for Computational Linguistics.

\bibitem[{Bertomeu et~al.(2006)Bertomeu, Uszkoreit, Frank, Krieger, and
  J{\"o}rg}]{Bertomeu06}
N{\'u}ria Bertomeu, Hans Uszkoreit, Anette Frank, Hans-Ulrich Krieger, and
  Brigitte J{\"o}rg. 2006.
\newblock Contextual phenomena and thematic relations in database qa dialogues:
  results from a wizard-of-oz experiment.
\newblock In \emph{Proceedings of the Interactive Question Answering Workshop
  at HLT-NAACL 2006}, pages 1--8. Association for Computational Linguistics.

\bibitem[{Chai and Jin(2004)}]{Chai04}
Joyce~Y. Chai and Rong Jin. 2004.
\newblock Discourse structure for context question answering.
\newblock In \emph{HLT-NAACL 2004 Workshop on Pragmatics in Question
  Answering}.

\bibitem[{Choi et~al.(2018)Choi, He, Iyyer, Yatskar, Yih, Choi, Liang, and
  Zettlemoyer}]{Eunsol18}
Eunsol Choi, He~He, Mohit Iyyer, Mark Yatskar, Wen-tau Yih, Yejin Choi, Percy
  Liang, and Luke Zettlemoyer. 2018.
\newblock Quac: Question answering in context.
\newblock In \emph{Proceedings of the 2018 Conference on Empirical Methods in
  Natural Language Processing}, pages 2174--2184. Association for Computational
  Linguistics.

\bibitem[{Dahl et~al.(1994)Dahl, Bates, Brown, Fisher, Hunicke-Smith, Pallett,
  Pao, Rudnicky, and Shriberg}]{Dahl94}
Deborah~A. Dahl, Madeleine Bates, Michael Brown, William Fisher, Kate
  Hunicke-Smith, David Pallett, Christine Pao, Alexander Rudnicky, and
  Elizabeth Shriberg. 1994.
\newblock Expanding the scope of the atis task: The atis-3 corpus.
\newblock In \emph{Proceedings of the Workshop on Human Language Technology},
  HLT '94, Stroudsburg, PA, USA. Association for Computational Linguistics.

\bibitem[{Dong and Lapata(2016)}]{dong16}
Li~Dong and Mirella Lapata. 2016.
\newblock Language to logical form with neural attention.
\newblock In \emph{Proceedings of the 54th Annual Meeting of the Association
  for Computational Linguistics, {ACL} 2016, August 7-12, 2016, Berlin,
  Germany, Volume 1: Long Papers}.

\bibitem[{Dong and Lapata(2018)}]{dong18}
Li~Dong and Mirella Lapata. 2018.
\newblock \href {http://aclweb.org/anthology/P18-1068} {Coarse-to-fine decoding
  for neural semantic parsing}.
\newblock In \emph{Proceedings of the 56th Annual Meeting of the Association
  for Computational Linguistics (Volume 1: Long Papers)}, pages 731--742.
  Association for Computational Linguistics.

\bibitem[{Finegan-Dollak et~al.(2018)Finegan-Dollak, Kummerfeld, Zhang,
  Ramanathan, Sadasivam, Zhang, and Radev}]{cathy18}
Catherine Finegan-Dollak, Jonathan~K. Kummerfeld, Li~Zhang, Karthik
  Ramanathan~Dhanalakshmi Ramanathan, Sesh Sadasivam, Rui Zhang, and Dragomir
  Radev. 2018.
\newblock Improving text-to-sql evaluation methodology.
\newblock In \emph{ACL 2018}. Association for Computational Linguistics.

\bibitem[{Frank(2013)}]{Frank13}
Stefan~L. Frank. 2013.
\newblock Uncertainty reduction as a measure of cognitive load in sentence
  comprehension.
\newblock \emph{Topics in cognitive science}, 5 3:475--94.

\bibitem[{Hale(2006)}]{Hale06}
John Hale. 2006.
\newblock Uncertainty about the rest of the sentence.
\newblock \emph{Cognitive science}, 30 4:643--72.

\bibitem[{Hemphill et~al.(1990)Hemphill, Godfrey, and Doddington}]{Hemphill90}
Charles~T. Hemphill, John~J. Godfrey, and George~R. Doddington. 1990.
\newblock \href {http://aclweb.org/anthology/H90-1021} {The atis spoken
  language systems pilot corpus}.
\newblock In \emph{Speech and Natural Language: Proceedings of a Workshop Held
  at Hidden Valley, Pennsylvania, June 24-27,1990}.

\bibitem[{Henderson et~al.(2014)Henderson, Thomson, and
  Williams}]{Henderson2014}
Matthew Henderson, Blaise Thomson, and Jason~D. Williams. 2014.
\newblock The second dialog state tracking challenge.
\newblock In \emph{SIGDIAL Conference}.

\bibitem[{Iyer et~al.(2017)Iyer, Konstas, Cheung, Krishnamurthy, and
  Zettlemoyer}]{iyer17}
Srinivasan Iyer, Ioannis Konstas, Alvin Cheung, Jayant Krishnamurthy, and Luke
  Zettlemoyer. 2017.
\newblock Learning a neural semantic parser from user feedback.
\newblock \emph{CoRR}, abs/1704.08760.

\bibitem[{Iyyer et~al.(2017)Iyyer, Yih, and Chang}]{Iyyer17}
Mohit Iyyer, Wen-tau Yih, and Ming-Wei Chang. 2017.
\newblock Search-based neural structured learning for sequential question
  answering.
\newblock In \emph{Proceedings of the 55th Annual Meeting of the Association
  for Computational Linguistics (Volume 1: Long Papers)}, pages 1821--1831.
  Association for Computational Linguistics.

\bibitem[{Kato et~al.(2004)Kato, Fukumoto, Masui, and
  Kando}]{Kato2004HandlingIA}
Tsuneaki Kato, Jun'ichi Fukumoto, Fumito Masui, and Noriko Kando. 2004.
\newblock Handling information access dialogue through qa technologies-a novel
  challenge for open-domain question answering.
\newblock In \emph{Proceedings of the Workshop on Pragmatics of Question
  Answering at HLT-NAACL 2004}.

\bibitem[{Levy(2008)}]{Levy08}
Roger~P. Levy. 2008.
\newblock Expectation-based syntactic comprehension.
\newblock \emph{Cognition}, 106:1126--1177.

\bibitem[{Li and Jagadish(2014)}]{li2014constructing}
Fei Li and HV~Jagadish. 2014.
\newblock Constructing an interactive natural language interface for relational
  databases.
\newblock \emph{VLDB}.

\bibitem[{Long et~al.(2016)Long, Pasupat, and Liang}]{long16}
Reginald Long, Panupong Pasupat, and Percy Liang. 2016.
\newblock Simpler context-dependent logical forms via model projections.
\newblock In \emph{Proceedings of the 54th Annual Meeting of the Association
  for Computational Linguistics (Volume 1: Long Papers)}, pages 1456--1465.
  Association for Computational Linguistics.

\bibitem[{Miller et~al.(1996)Miller, Stallard, Bobrow, and Schwartz}]{Miller96}
Scott Miller, David Stallard, Robert~J. Bobrow, and Richard~M. Schwartz. 1996.
\newblock A fully statistical approach to natural language interfaces.
\newblock In \emph{ACL}.

\bibitem[{Mrk{\v{s}}i{\'{c}} et~al.(2017)Mrk{\v{s}}i{\'{c}},
  {\'O}~S{\'e}aghdha, Wen, Thomson, and Young}]{mrksic17}
Nikola Mrk{\v{s}}i{\'{c}}, Diarmuid {\'O}~S{\'e}aghdha, Tsung-Hsien Wen, Blaise
  Thomson, and Steve Young. 2017.
\newblock \href {https://doi.org/10.18653/v1/P17-1163} {Neural belief tracker:
  Data-driven dialogue state tracking}.
\newblock In \emph{Proceedings of the 55th Annual Meeting of the Association
  for Computational Linguistics (Volume 1: Long Papers)}, pages 1777--1788.
  Association for Computational Linguistics.

\bibitem[{Pasupat and Liang(2015)}]{pasupat2015compositional}
Panupong Pasupat and Percy Liang. 2015.
\newblock Compositional semantic parsing on semi-structured tables.
\newblock In \emph{Proceedings of the 53rd Annual Meeting of the Association
  for Computational Linguistics and the 7th International Joint Conference on
  Natural Language Processing of the Asian Federation of Natural Language
  Processing, {ACL} 2015, July 26-31, 2015, Beijing, China, Volume 1: Long
  Papers}, pages 1470--1480.

\bibitem[{Pennington et~al.(2014)Pennington, Socher, and
  Manning}]{pennington14}
Jeffrey Pennington, Richard Socher, and Christopher~D. Manning. 2014.
\newblock Glove: Global vectors for word representation.
\newblock In \emph{{EMNLP}}, pages 1532--1543. {ACL}.

\bibitem[{Popescu et~al.(2003)Popescu, Etzioni, and Kautz}]{popescu2003towards}
Ana-Maria Popescu, Oren Etzioni, and Henry Kautz. 2003.
\newblock Towards a theory of natural language interfaces to databases.
\newblock In \emph{Proceedings of the 8th international conference on
  Intelligent user interfaces}, pages 149--157. ACM.

\bibitem[{Reddy et~al.(2018)Reddy, Chen, and Manning}]{Reddy18}
Siva Reddy, Danqi Chen, and Christopher~D. Manning. 2018.
\newblock Coqa: A conversational question answering challenge.
\newblock \emph{CoRR}, abs/1808.07042.

\bibitem[{Shi et~al.(2018)Shi, Tatwawadi, Chakrabarti, Mao, Polozov, and
  Chen}]{shi2018incsql}
Tianze Shi, Kedar Tatwawadi, Kaushik Chakrabarti, Yi~Mao, Oleksandr Polozov,
  and Weizhu Chen. 2018.
\newblock Incsql: Training incremental text-to-sql parsers with
  non-deterministic oracles.
\newblock \emph{arXiv preprint arXiv:1809.05054}.

\bibitem[{Suhr et~al.(2018)Suhr, Iyer, and Artzi}]{Suhr:18context}
Alane Suhr, Srinivasan Iyer, and Yoav Artzi. 2018.
\newblock \href {http://aclweb.org/anthology/N18-1203} {Learning to map
  context-dependent sentences to executable formal queries}.
\newblock In \emph{Proceedings of the Conference of the North American Chapter
  of the Association for Computational Linguistics: Human Language
  Technologies}, pages 2238--2249. Association for Computational Linguistics.

\bibitem[{Wang et~al.(2018)Wang, Huang, Polozov, Brockschmidt, and
  Singh}]{2018executionguided}
Chenglong Wang, Po{-}Sen Huang, Alex Polozov, Marc Brockschmidt, and Rishabh
  Singh. 2018.
\newblock Execution-guided neural program decoding.
\newblock In \emph{ICML workshop on Neural Abstract Machines and Program
  Induction v2 (NAMPI)}.

\bibitem[{Xu et~al.(2017)Xu, Liu, and Song}]{Xu2017}
Xiaojun Xu, Chang Liu, and Dawn Song. 2017.
\newblock Sqlnet: Generating structured queries from natural language without
  reinforcement learning.
\newblock \emph{arXiv preprint arXiv:1711.04436}.

\bibitem[{Yu et~al.(2018{\natexlab{a}})Yu, Li, Zhang, Zhang, and Radev}]{Yu18}
Tao Yu, Zifan Li, Zilin Zhang, Rui Zhang, and Dragomir Radev.
  2018{\natexlab{a}}.
\newblock Typesql: Knowledge-based type-aware neural text-to-sql generation.
\newblock In \emph{Proceedings of NAACL}. Association for Computational
  Linguistics.

\bibitem[{Yu et~al.(2018{\natexlab{b}})Yu, Yasunaga, Yang, Zhang, Wang, Li, and
  Radev}]{Yu&al.18.emnlp.syntax}
Tao Yu, Michihiro Yasunaga, Kai Yang, Rui Zhang, Dongxu Wang, Zifan Li, and
  Dragomir Radev. 2018{\natexlab{b}}.
\newblock Syntaxsqlnet: Syntax tree networks for complex and cross-domain
  text-to-sql task.
\newblock In \emph{Proceedings of EMNLP}. Association for Computational
  Linguistics.

\bibitem[{Yu et~al.(2018{\natexlab{c}})Yu, Zhang, Yang, Yasunaga, Wang, Li, Ma,
  Li, Yao, Roman, Zhang, and Radev}]{Yu18emnlp}
Tao Yu, Rui Zhang, Kai Yang, Michihiro Yasunaga, Dongxu Wang, Zifan Li, James
  Ma, Irene Li, Qingning Yao, Shanelle Roman, Zilin Zhang, and Dragomir Radev.
  2018{\natexlab{c}}.
\newblock Spider: A large-scale human-labeled dataset for complex and
  cross-domain semantic parsing and text-to-sql task.
\newblock In \emph{EMNLP}.

\bibitem[{Zelle and Mooney(1996)}]{zelle96}
John~M. Zelle and Raymond~J. Mooney. 1996.
\newblock Learning to parse database queries using inductive logic programming.
\newblock In \emph{AAAI/IAAI}, pages 1050--1055, Portland, OR. AAAI Press/MIT
  Press.

\bibitem[{Zettlemoyer and Collins(2007)}]{D07-1071}
Luke Zettlemoyer and Michael Collins. 2007.
\newblock \href {http://aclweb.org/anthology/D07-1071} {Online learning of
  relaxed ccg grammars for parsing to logical form}.
\newblock In \emph{Proceedings of the 2007 Joint Conference on Empirical
  Methods in Natural Language Processing and Computational Natural Language
  Learning (EMNLP-CoNLL)}.

\bibitem[{Zettlemoyer and Collins(2005)}]{Zettlemoyer05}
Luke~S. Zettlemoyer and Michael Collins. 2005.
\newblock Learning to map sentences to logical form: Structured classification
  with probabilistic categorial grammars.
\newblock \emph{UAI}.

\bibitem[{Zettlemoyer and Collins(2009)}]{Zettlemoyer09}
Luke~S. Zettlemoyer and Michael Collins. 2009.
\newblock Learning context-dependent mappings from sentences to logical form.
\newblock In \emph{ACL/IJCNLP}.

\bibitem[{Zhong et~al.(2017)Zhong, Xiong, and Socher}]{Zhong2017}
Victor Zhong, Caiming Xiong, and Richard Socher. 2017.
\newblock Seq2sql: Generating structured queries from natural language using
  reinforcement learning.
\newblock \emph{CoRR}, abs/1709.00103.

\bibitem[{Zhong et~al.(2018)Zhong, Xiong, and Socher}]{zhong2018global}
Victor Zhong, Caiming Xiong, and Richard Socher. 2018.
\newblock Global-locally self-attentive encoder for dialogue state tracking.
\newblock In \emph{ACL}.

\end{thebibliography}
